\documentclass{ieeeaccess}
\usepackage{comment}
\usepackage{cite}
\usepackage{amsmath,amssymb,amsfonts}
\usepackage{algorithmic}
\usepackage{graphicx}
\usepackage{textcomp}
\usepackage{hyperref}
\usepackage{booktabs} 
\usepackage{amsmath}

\def\BibTeX{{\rm B\kern-.05em{\sc i\kern-.025em b}\kern-.08em
    T\kern-.1667em\lower.7ex\hbox{E}\kern-.125emX}}

\begin{document}
\history{Date of publication xxxx 00, 0000, date of current version xxxx 00, 0000.}
\doi{XX.XXXX/ACCESS.20XX.DOI}

\title{Enhanced LFTSformer: A Novel Long-Term Financial Time Series Prediction Model Using Advanced Feature Engineering and the DS Encoder Informer Architecture}

\author
{
\uppercase{Jianan Zhang}  
\authorrefmark{1} \IEEEmembership{Student Member, IEEE},
\uppercase{Hongyi Duan}   
\authorrefmark{2,3} \IEEEmembership{Student Member, IEEE}.
}

\address[1]{School of Mathematica, Shanghai University of Finance and Economics, Yangpu District, Shanghai 200437 China}

\address[2]{Faculty of Electronic and Information Engineering, Xi'an Jiaotong University, Xi'an, Shannxi  710049 China }

\address[3]{School of Computing Information System, The University of Melbourne, Melbourne, VIC  3010 Australia} 
\tfootnote{JIANAN ZHANG and HONGYI DUAN Contributed to the work equally and should be considered co-first authors}
\corresp{Corresponding author: Hongyi Duan (e-mail: Dann\_Hiroaki@ieee.org).}

\begin{abstract}
This study presents a groundbreaking model for forecasting long-term financial time series, termed the Enhanced LFTSformer. The model distinguishes itself through several significant innovations:
\begin{enumerate}
\item \textbf{VMD-MIC+FE Feature Engineering: }The incorporation of sophisticated feature engineering techniques, specifically through the integration of Variational Mode Decomposition (VMD), Maximal Information Coefficient (MIC), and feature engineering (FE) methods, enables comprehensive perception and extraction of deep-level features from complex and variable financial datasets.
\item \textbf{DS Encoder Informer: }The architecture of the original Informer has been modified by adopting a Stacked Informer structure in the encoder, and an innovative introduction of a multi-head decentralized sparse attention mechanism, referred to as the Distributed Informer. This modification has led to a reduction in the number of attention blocks, thereby enhancing both the training accuracy and speed.
\item \textbf{$\text{GC}$ Enhanced Adam \& Dynamic Loss Function: }The deployment of a Gradient Clipping-enhanced Adam optimization algorithm and a dynamic loss function represents a pioneering approach within the domain of financial time series prediction. This novel methodology optimizes model performance and adapts more dynamically to evolving data patterns.
\end{enumerate}
Systematic experimentation on a range of benchmark stock market datasets demonstrates that the Enhanced LFTSformer outperforms traditional machine learning models and other Informer-based architectures in terms of prediction accuracy, adaptability, and generality. Furthermore, the paper identifies potential avenues for future enhancements, with a particular focus on the identification and quantification of pivotal impacting events and news. This is aimed at further refining the predictive efficacy of the model.
\end{abstract}

\begin{keywords}
Feature Engineering, Financial Time Series, Integrated Models, Stacked Informants, Variational Mode Decomposition.
\end{keywords}

\titlepgskip=-15pt

\maketitle

\section{Introduction}
\subsection{Research Background}
The financial market constitutes a pivotal foundation for the sustenance of global economic endeavors, with its operations and dynamical shifts intricately intertwined with a confluence of complex determinants, encompassing economic structuration, seasonal variabilities, and the international milieu\cite{1}\cite{2}. Concomitant with economic progression and the burgeonment of financial markets, the adoption of time series analysis has emerged as an indispensable instrumentality within the finance domain.\cite{3} This analytical paradigm has markedly enhanced the comprehension of market dynamics, augmented the intelligence of decision-making processes, and facilitated the prognostication of investment returns.\cite{2}\cite{4} Consequently, time series analysis has garnered extensive scholarly interest, engendering a copious corpus of research findings and advancing the academic discourse on financial market analysis.\cite{5}\cite{6}

The domain of long-term time series forecasting, increasingly pivotal across an array of disciplines, encapsulates several quintessential applications including the temporal allocation mechanisms pivotal for wind energy integration\cite{7}, the granular dissection of protracted energy consumption patterns within architectural edifices\cite{8}, and the nuanced prognostication of load dynamics within thermal infrastructures\cite{9}, among others. In marked divergence from the traditional paradigms of time series forecasting, the arena of financial time series forecasting is imbued with pronounced complexity and a heightened degree of unpredictability. This complexity is particularly accentuated in scenarios where prognostications are extended towards long-term horizons or when the forecasts are subjected to unforeseen exogenous perturbations\cite{10}\cite{11}. 

Research on financial time series forecasting has continuously evolved since the 1950s, despite numerous challenges. Initially, the field predominantly employed pure statistical and mathematical methodologies. Over time, it incorporated machine learning and deep learning techniques. Contemporary research focuses on the refinement of these models through the integration of advanced attention mechanisms and sophisticated feature engineering strategies, representing the cutting edge in financial time series analysis.

\subsection{Financial Time-Series Prediction: A Review}
\subsubsection{Classical Theories in Financial Time Series Analysis}
Within the domains of finance and macroeconomics, the inaugural application of mathematical models, anchored in the principles of econometrics and statistical theory, has pioneered the analytical and prognostic assessment of time series data. Notably, Robert Engel's seminal introduction of the Autoregressive Conditional Heteroskedasticity (ACH) framework, while examining the volatility of UK inflation rates, has emerged as a significant landmark in the evolution of financial market volatility forecasting methodologies\cite{16}. This foundational work was subsequently advanced by Tim Boleslef, who refined it into the Generalized Autoregressive Conditional Heteroskedasticity (GARCH) model, thereby significantly broadening the analytical precision and applicability of this approach within the realm of financial time series analysis\cite{17}. Complementing these developments, David A. Shea's exploration into the volatility of five major foreign exchange rates, through the employment of a sophisticated statistical model that accounts for time-variant mean and variance, has further augmented the repertoire of financial market forecasting methodologies\cite{18}. Part of this era's research is distinguished by the strategic amalgamation of diverse financial time series datasets, fostering a multidimensional analytical perspective\cite{19}.

In the 21st century's dynamic financial landscape, marked by burgeoning economic activities and market expansion, there has been a proliferation of statistical models for financial time series forecasting. The most remarkable is RG Brown's introduction of the ARIMA model in 2004\cite{22} has catalyzed its adoption for predicting stock and futures market trends, alongside its variants like AR, VAR, ARDL, ARCH, GARCH, and MIDAS\cite{23}. These models excel in linear stock price scenarios, effectively assessing prediction risks\cite{24} and forecasting returns\cite{25}. Nonetheless, their predictive power for highly volatile stocks influenced by multifarious factors remains debated. To address this challenge, a large number of fundamental machine learning models and methods have been progressively incorporated into the forecasting of financial and stock time series under the rise of the artificial intelligence field.

\subsubsection{Introduction of machine learning to financial time series prediction}

The utilization of machine learning techniques in forecasting financial time series marks the advent of sophisticated analytical methodologies. The introduction of temporal difference learning by Sutton signals a pivotal shift, utilizing historical data to enhance predictive accuracy while maximizing computational resourcefulness\cite{26}. The deployment of genetic algorithms by Mahfoud and Mani, along with Kim and Han, has refined the processes of network training and feature selection\cite{27}\cite{28}. Additionally, the integration of Support Vector Machines (SVM) by Trafalis and Ince, enriched by Lee, has notably augmented the precision of stock trend predictions, underscoring the utility of composite feature selection techniques\cite{29}\cite{30}.

Deep learning, a significant evolution within machine learning, has revolutionized financial forecasting with marked improvements in precision and the ability to generate deep insights. Chen et al.’s employment of LSTM for the prediction of returns in China’s stock market illustrates a considerable leap over traditional methodologies, adeptly navigating market unpredictability\cite{31}. The CNN-GA model developed by Li et al. for short-term pork price prediction showcases deep learning's adaptability, yielding high precision and underscoring its value in the analysis of commodity markets\cite{32}. Ding et al.’s model for event-driven forecasting, which leverages deep convolutional neural networks, offers an intricate analysis of how events influence stock prices, providing a profound augmentation over existing baselines\cite{33}.

\subsubsection{Improvement of machine learning based financial time series Prediction}

In the context of the rapidly growing volume of data in today's financial markets, machine learning and deep learning algorithms have demonstrated superior performance in time series forecasting compared to traditional forecasting methods. However, these advanced algorithms are not without limitations\cite{34}. A particularly prominent challenge is that financial time series data is too complex and variable, resulting in an inability to adequately extract salient features\cite{35}. In addition, the fidelity of predictions tends to decrease over time, with RNNs and LSTMs being examples of decreasing predictive effectiveness\cite{37}. To address these challenges, a large body of academic work has been devoted to the design of sophisticated feature engineering, as well as the introduction and optimisation of attentional mechanisms suitable for long time-series forecasting, which has yielded promising results in the field of financial forecasting.

Our research focuses primarily on attention mechanisms and the Informer model. Since Vaswani et al. established the fully attention-based Transformer model in 2016, which does not rely on recurrence and convolution\cite{38}, it has triggered a paradigm shift in the fields of machine translation and computer vision\cite{38*}\cite{38**}, and has been widely recognized for its outstanding performance in areas such as time series prediction\cite{38***}. Following this, Zhou et al\cite{39}. introduced Informer, aimed at addressing the complexities of long-sequence time series forecasting (LSTF) by incorporating an efficient attention mechanism that notably diminishes time complexity and memory consumption while preserving long-term dependencies with remarkable precision. This advancement, alongside further enhancements and innovations in the Transformer architecture, was promptly integrated into the domain of long-term financial time series forecasting. 

As for Feature Engineering and other aspects, it merits attention that additional significant optimizations and inquiries into long-term time series forecasting have been explored, including stacking-based ensemble methodologies\cite{46}, and innovations in network structure fusion\cite{47}\cite{48}, grounded in comprehensive parameter pruning strategies. Additionally, the decomposition-reconstruction approach, utilizing Variational Mode Decomposition (VMD) and Empirical Mode Decomposition (EMD)\cite{49}\cite{50}\cite{51}, has been identified as another pivotal research area. These optimization techniques, serving as integral components of our study, will also manifest in the context of feature engineering, thereby augmenting the efficacy and applicability of our research.

\subsection{Our Research Centre and Article Structure}
In summary, in the context of adequate feature extraction and accurate prediction for long time series, we combine the latest advances with the introduction of the pioneering Enhanced LFTS Informer, which combines the advantages of VMD-MIC/FE in the feature engineering stage, and through the mechanisms of Encoder Stacking/Decentralisation, coupled with the $\text{GC}$-enhanced Adam Optimisation as well as the Dynamic The introduction of Loss Function achieves excellent results in long term financial time series prediction (this time stocks are selected).

The rest of this article will be developed and discussed in the following order
\begin{enumerate}
\item \textbf{Section \ref{2}}: A review of the progress and shortcomings of informer-based modules in financial time series and other time series forecasting in recent years.
\item \textbf{Section \ref{3}}: Describes the overall architecture of the article's research, including data preprocessing before feature engineering, feature engineering methods, DS Encoder Informer architecture, etc.
\item \textbf{Section \ref{4}}: Demonstrates the process of feature engineering and results of ablation experiments / comparison experiments / stability experiments
\item \textbf{Section \ref{5}}: Summarize the conclusions of this article and critically proposes potential improvement directions for this method.
\end{enumerate}

\section{Related Works} \label{2}
\subsection{Informer-based Model in Financial Time Series Prediction}
In the current research field of financial time series forecasting, studies based on Informer models are not only scarce in number, but also of varying quality. Despite these challenges, some representative studies demonstrate the potential and effectiveness of Informer models when dealing with complex financial data. By analysing these studies, we can gain a deeper understanding of the application of Informer models in stock price forecasting and market trend analysis.

First, one category of research focuses on improving the accuracy of stock price forecasting. For example, Ren et al. developed a hybrid model combining Encoder Forest and Informer, which significantly improves the forecasting ability through the decomposition and reconstruction approach \cite{52}.Ding et al. then optimised intraday stock price forecasting using the Informer model, demonstrating its superiority in capturing long-range dependencies \cite{54}. These studies highlight the ability of Informer models in accurately capturing market dynamics.

Secondly, another class of studies focuses on the improvement of long-term forecasting.Liu et al. employed the Informer model optimised by the PSO algorithm to provide more accurate forecasts for long-term stock price trends, overcoming the shortcomings of the traditional model in long series forecasting\cite{53}. In addition, Sababipour Asl used the Informer model to improve the prediction accuracy of stock market volatility, demonstrating its potential application in financial risk management\cite{55}.

Finally, there are also studies focusing on the versatility and integrated application of models. For example, Abdulsahib and Ghaderi demonstrated the advantages of the Informer encoder in capturing global and local market dynamics, proposing a new financial market prediction model\cite{56}. In addition, Fei Xiong and Yuhao Feng's master's thesis investigated quantitative trading strategies based on the Informer mechanism, verifying the applicability and effectiveness of the model in real financial trading \cite{57}\cite{58}.

\subsection{Informer-based Model in Other Time Series Prediction}
Informer-based models have not been widely studied in the field of financial time series. Therefore, for analysis and reference, several representative Informer-based models from other time-series prediction domains are selected for this review. These models mainly focus on three aspects, namely structural improvement, performance optimisation and application-specific customisation, in order to improve prediction accuracy, processing power and application adaptability. Thus, they provide possible insights and directions for financial time series prediction

In terms of the underlying structural optimisation of the model structure, the MSRN-Informer model enhances the ability to extract features from time-series data by integrating the multi-scale structure and residual network, reduces information loss, and effectively improves the limitations of the traditional deep network in processing time-series data \cite{62}. In addition, the Enformer model employs a simplified Transformer structure that uses only the encoder and omits the decoder, which reduces the computational complexity by introducing a sparse periodic attention mechanism, and at the same time enhances the model's ability to capture the periodic dependence of the sequence \cite{67}

In terms of performance optimisation, the MTS-Informer model enhances the model's ability to handle complex multivariate data by introducing causal convolution and a sparse self-attention mechanism, and improves the handling of unstable sequences\cite{60}. Literature \cite{63} demonstrates the application of Informer in the prediction of motor bearing vibration, which optimises the temporal and spatial complexity and improves the prediction accuracy and reliability of the model by employing ProbSparse self-attention and self-attention distillation techniques.

In terms of customised tuning for specific application scenarios, the improved stacked Informer network is designed for power line fault prediction, optimising the network's generalisation capability and training speed by incorporating gradient concentration techniques \cite{64}. Whereas the model in literature \cite{65} combines the integration of Empirical Modal Decomposition (EEMD) with Particle Swarm Optimisation algorithm (PSO) tuning of Informer, specifically designed for multi-step prediction of building energy consumption to cope with non-linear and non-smooth energy consumption data.

\section{Methodology} \label{3}

\subsection{Data Preprocessing}\label{1}
\subsubsection{Raw Data}
In this study, we adopt stocks, the most common and widespread financial time series, as the object of study. Based on this, we collect the stocks of 22 large enterprises listed on the two major stock exchanges in China (Shanghai Stock Exchange and Shenzhen Stock Exchange) as the research object, and the basic information is shown in Table \ref{tab:stock_data}.

\begin{table}[ht]
\centering
\caption{Data Collected and Used in This Study}
\renewcommand{\arraystretch}{1.3}
\label{tab:stock_data}
\begin{tabular}{|c|c|c|c|c|}
\hline
\textbf{Stock Code} & \textbf{Volume} & \textbf{Interval} & \textbf{Start Date} & \textbf{End Date} \\
\hline
SHA: 601398 & 190225 & 30min & 2006/10/30 & 2023/03/17 \\
SHE: 002415 & 146833 & 30min & 2010/05/28 & 2023/03/21 \\
SHA: 600007 & 269429 & 30min & 1999/10/28 & 2023/03/21 \\
SHA: 601628 & 188545 & 30min & 2007/01/10 & 2023/03/21 \\
SHA: 601857 & 179041 & 30min & 2007/11/06 & 2023/03/21 \\
SHA: 600654 & 244903 & 30min & 1999/10/28 & 2023/03/21 \\
SHA: 600004 & 229681 & 30min & 2003/04/29 & 2023/03/21 \\
SHE: 300059 & 148609 & 30min & 2010/03/19 & 2023/03/21 \\
SHA: 600006 & 266793 & 30min & 1999/10/28 & 2023/03/21 \\
SHE: 000004 & 255569 & 30min & 1999/10/28 & 2023/03/21 \\
SHA: 600009 & 269384 & 30min & 1999/10/28 & 2023/03/21 \\
SHE: 000938 & 248423 & 30min & 1999/11/11 & 2023/03/21 \\
SHE: 600653 & 258001 & 30min & 1999/10/28 & 2023/03/21 \\
SHA: 600601 & 268896 & 30min & 1999/10/28 & 2023/03/21 \\
SHA: 601800 & 128689 & 30min & 2012/03/09 & 2023/03/21 \\
SHA: 600000 & 265010 & 30min & 1999/12/07 & 2023/03/17 \\
SHE: 000001 & 262608 & 30min & 1999/12/21 & 2023/03/21 \\
SHA: 600602 & 255553 & 30min & 1999/10/28 & 2023/03/21 \\
SHE: 000089 & 268902 & 30min & 1999/10/27 & 2023/03/21 \\
SHA: 600028 & 249217 & 30min & 2001/08/08 & 2023/03/21 \\
SHE: 000063 & 264673 & 30min & 1999/10/27 & 2023/03/21 \\
SHA: 600065 & 251770 & 30min & 1999/10/28 & 2023/03/21 \\
\hline
\end{tabular}
\end{table}

For the raw data collected, the basic paradigm and Attributes are shown in Table \ref{tab:stock_data_example}. As for the complete initial data, please refer to \href{https://github.com/DANNHIROAKI/SH-Stock-Price}{this link}. One thing to note is that all data related to transaction volume are in RMB.

\begin{table}[ht]
\centering
\caption{Stock Raw Data Format Example}
\renewcommand{\arraystretch}{1.3}
\label{tab:stock_data_example}
\begin{tabular}{|c|c|c|c|c|c|c|c|}
\hline
 \textbf{Time} & \textbf{Open} & \textbf{High} & \textbf{Low} & \textbf{Close} & \textbf{Volume} & \textbf{Amount} \\
\hline
 199910280950 & 6.3 & 6.33 & 6.3 & 6.32 & 62700 & 395321 \\
 199910280955 & 6.32 & 6.32 & 6.3 & 6.32 & 84400 & 532593 \\
 199910281000 & 6.31 & 6.33 & 6.31 & 6.31 & 37600 & 237544 \\
 199910281005 & 6.3	& 6.31 & 6.3  & 6.31 & 37700 & 237674 \\
 ············ & ···	& ··· & ···  & ···· & ····· & ······ \\
\hline
\end{tabular}
\end{table}

\subsubsection{Data Indexing}

To fully reflect the inherent patterns in the data, we extracted 29 features from the original dataset. This includes 8 basic indicators as shown in Table \ref{basic_features}. Building on the studies by Achelis \cite{66}, Kara et al. \cite{67}, and Guresen et al. \cite{68}, we identified an additional 21 advanced features for this research, as listed in Table \ref{features}. While the calculation methods for all advanced features are comprehensively discussed in Achelis \cite{66}.

\begin{table}[ht]
\centering
\caption{Basic Features}
\renewcommand{\arraystretch}{1.4}
\label{basic_features}
\begin{tabular}{cccc}
\hline  Feature name & Window size & Number of features \\
\hline  Open, Close, High, Low & 1 & 4 \\
 Weighted Price & 1 & 1 \\
 Volume(Stock), Volume(RMB) & 1 & 2 \\
 High - Low & 1 & 1\\
\hline
\end{tabular}
\end{table}

\begin{table}[ht]
\centering
\caption{Advanced Features}
\renewcommand{\arraystretch}{1.4}
\label{features}
\begin{tabular}{cccc}
\hline Feature name & Window size & Number of features \\
\hline 
Return & $1, \ldots, 6$ & 6 \\
Correlation MA5 and MA30 & 1 & 1 \\
Sum3, Sum5 & 5,3 & 2 \\
Sum5 - Sum3 & 1 & 1 \\
|Sum5 - Sum3| & 1 & 1 \\
$\sqrt{\text{(Sum5)}^2 - \text{(Sum3)}^2}$ & 1 & 1 \\
RSI(6, 14) & 6,14 & 2 \\
Rate of change & 9,14 & 2 \\
Williams R & 1 & 1 \\
ATR & 5,10 & 2 \\
CCI & 1 & 1 \\
DEMA & 1 & 1 \\
\hline
\end{tabular}
\end{table}

Due to space constraints, we only explain the following specific indicators here.

\begin{enumerate}
\item \textbf{Return: }Refers to returns at 6 different lag levels (from 1 to 6 lag length)
\item \textbf{Average True Range (ATR):} Calculated by the following formula:
\begin{equation}
\text{ART} = \mathrm{MA}\text{(TR,N)}
\end{equation}
Where $\mathrm{MA}(T R, N)$ represents the moving average of $T R$ over $N$ time units, i.e., the average price over the past $N$ time periods. The true range $T R_t$ for the $t^{th}$ time period is computed as:
\begin{equation}
\begin{aligned}
\text{TR}_t = \max \left\{ \left[ \begin{array}{c}
\text{High}_t \\
\text{High}_t  \\
\text{Close}_{t-1}
\end{array} \right] - 
\left[ \begin{array}{c}
\text{Low}_t \\
\text{Close}_{t-1} \\
\text{Low}_t
\end{array} \right] \right\}
\end{aligned}
\end{equation}
where High, Low, and Close are the highest, lowest, and closing prices respectively.

\item \textbf{Relative Strength Index (RSI):} Calculated as:
\begin{equation}
\text{RSI} = 100 - \cfrac{100}{1 + \cfrac{\mathrm{MA}(U, N)}{\mathrm{MA}(D, N)}}
\end{equation}
When the price increases, we have:
\begin{equation}
\text{U}_t = \text{Close}_t - \text{Close}_{t-1}, \quad \text{D}=0
\end{equation}
and when it decreases, we have:
\begin{equation}
\text{D}_t = \text{Close}_{t-1} - \text{Close}_t, \quad \text{U}=0
\end{equation}

\item \textbf{Price Return:} Calculated with the following formula:
\begin{equation}
\text{Return}_{i_t} = \text{Close}_t - \text{Close}_{t-i}
\end{equation}

\item \textbf{Rate of Change (ROC):} Calculated as:
\begin{equation}
\text{ROC}_{i_t} = 100 \cdot \left(\cfrac{\text{Close}_t}{\text{Close}_{t-i}} - 1\right)
\end{equation}

\item \textbf{Stochastic Oscillator:} Reflects the momentum effect in the Bitcoin price, and its calculation formula is as follows:
\begin{equation}
\text{K} = 100 \cdot \cfrac{\text{Close} - \text{Low}}{\text{High} - \text{Low}}
\end{equation}
$$
\downarrow
$$
\begin{equation}
\text{Stochastic oscillator} = \mathrm{MA}\text{(K, N)}
\end{equation}

\item \textbf{Commodity Channel Index (CCI):} Reflects the average price deviation, calculated as:
\begin{equation}
\begin{cases}
\text{TP} = \cfrac{\text{High} + \text{Low} + \text{Close}}{3} \\\\
\text{MC} = \mathrm{MA}(\text{Close}, \text{N}) \\\\
\text{MD} = \mathrm{MA}(\text{MC} - \text{Close}, \text{N})
\end{cases}
\end{equation}
$$
\downarrow
$$
\begin{equation}
\begin{aligned}
\text{CCI} & = \cfrac{\text{TP} - \text{MC}}{0.015\text{MD}} 
\end{aligned}
\end{equation}
\end{enumerate}

\subsubsection{Data Manipulation}
Given the heightened complexity inherent to stock prices, the natural logarithm of the terminal prices was extracted, followed by first-order differencing. The corresponding formula is presented as:

\begin{align}
\text{new}\: \text{close}_i=\log(\cfrac{\text{close}_i}{\text{close}_{i-1}})
\end{align}
To enhance the efficiency of our neural network model, we standardized the 29 indicators. The calculation formula is:
\begin{align}
\text{new}\: \text{value}_i=\cfrac{\text{value}_i-\text{E(value)}}{\sqrt{\text{D(value)}}}
\end{align}
Where \( \text{E(value)} \) and \( \text{D(value)} \) represent the expected value and variance of the indicator, respectively.

\subsection{Feature Engineering}

\subsubsection{VMD Model}

Variance Mode Decomposition (VMD) is a sophisticated data decomposition technique, facilitates the transformation of time-domain financial sequences into the frequency domain\cite{69}. This transformation yields a set of \( K \) intrinsic mode functions (IMFs). The foundational step in this procedure entails formulating the variational constraint equation:

\begin{equation}
\left\{\begin{array}{l}
\min \left\{\displaystyle\sum\limits_{\text{K=1}}^\text{K}\left\|\partial_t\left[\left(\delta(t)+\cfrac{i}{\pi t}\right) * u_{\text{K}}(t)\right] e^{-i w_{\text{K}} t}\right\|_2^2\right\} \\\\
\text { s.t. } \displaystyle\sum\limits_{\text{K=1}}^\text{K} u_{\text{K}}=x(t)
\end{array}\right.
\end{equation}

Here, \( x(t) \) denotes the financial data sequence. The terms \( w_{\text{K}} \) and \( u_{\text{K}} \) respectively represent the central frequency and band components of the \( k \)-th IMF. The impulse function is represented by \( \delta(t) \), while \( \partial t \) indicates the derivative with respect to time. The convolution operation (*) signifies the convolution computation in this context.

For a more tractable representation, the variational constraint equation is restructured using the Lagrange function \( \lambda(t) \) and the penalty factor \( \alpha \):

\begin{equation}
\begin{aligned}
\begin{cases}
\text{TDR} &= \alpha \displaystyle\sum\limits_{{\text{K}}=1}^{\text{K}} \left\| \partial_t \left[ \left( \delta(t) + \cfrac{i}{\pi t} \right) * u_{\text{K}}(t) \right] e^{-i w_{\text{K}} t} \right\|_2^2 \\\\
\text{DAE} &= \left\| x(t) - \displaystyle\sum\limits_{{\text{K}}=1}^{\text{K}} u_{\text{K}}(t) \right\|_2^2 \\\\
\text{LM}  &= \left\langle \lambda(t), x(t) - \displaystyle\sum\limits_{{\text{K}}=1}^{\text{K}} u_{\text{K}}(t) \right\rangle
\end{cases}
\end{aligned}
\end{equation}
$$
\downarrow
$$
\begin{equation}
\text{L}=(\{u_{\text{K}}\},\{\omega{}_{\text{K}}\},\lambda) = \text{TDR}+\text{DAE}+\text{LM}
\end{equation}

This reformulation alleviates the constraints, rendering the problem more manageable. The optimal solution is then determined using the Alternating Direction Method of Multipliers (ADMM), a renowned optimization technique. The iterative steps are as follows:
\begin{equation}
\begin{gathered}
\hat{u}_{\text{K}}^{n+1}(w)=\cfrac{\hat{x}(w)-\displaystyle\sum\limits_{i \neq {\text{K}}} \hat{u}_j(w)+\cfrac{\hat{\lambda}(w)}{2}}{1+2 \alpha\left(w-w_{\text{K}}\right)^2} \\
\end{gathered}
\end{equation}
$$
\downarrow
$$
\begin{equation}
\begin{gathered}
w_{\text{K}}^{n+1}=\cfrac{\displaystyle\int\limits_0^{\infty} w\left|\hat{u}_{\text{K}}^{n+1}(w)\right|^2 d w}{\displaystyle\int\limits_0^{\infty}\left|\hat{u}_{\text{K}}^{n+1}(w)\right|^2 d w}
\end{gathered}
\end{equation}

Following this sequence, the primary financial data sequence is decomposed into \( K \) distinct sub-series.

\subsubsection{VMD-MIC Framework}
While VMD is underpinned by a robust mathematical framework and adept at disentangling complex signal components, it's constrained by the need for predetermined decomposition parameters. This requirement can compromise the fidelity of the decomposition. To address these limitations, this study proposes an amalgamation of VMD with the Mutual Information Coefficient (MIC)\cite{70}. This integration determines the optimal number of decompositions, \( {{K}} \), enhancing the decomposition's adaptability.

The Mutual Information Criterion (MIC) is a statistical measure used to quantify the dependency between two variables, indicating the amount of information shared between them. As for the Calculation of the Mutual Information Criterion (MIC), when given two variables, \(X\) and \(Y\), their mutual information \(I(X; Y)\) is defined as:
\begin{equation}
\text{I}(\text{X}; \text{Y}) = \displaystyle\sum\limits_{x \in \text{X}} \displaystyle\sum\limits_{y \in \text{Y}} p(x, y) \log \cfrac{p(x, y)}{p(x)p(y)}
\end{equation}
Here, \(p(x, y)\) represents the joint probability distribution of \(X\) and \(Y\), while \(p(x)\) and \(p(y)\) are the marginal probability distributions of \(X\) and \(Y\), respectively.

The Mutual Information Criterion (MIC) is a normalized form of mutual information, calculated as follows:
\begin{equation}
\text{MIC}(\text{X}; \text{Y}) = \cfrac{\text{I}(\text{X}; \text{Y})}{\sqrt{H(\text{X})H(\text{Y})}}
\end{equation}
In this equation, \(H(X)\) and \(H(Y)\) represent the entropy of \(X\) and \(Y\) respectively, used to normalize the mutual information such that the values of MIC range from 0 to 1.

In the context of Variational Mode Decomposition (VMD), the Mutual Information Criterion (MIC) serves as a robust measure for quantifying the accuracy and effectiveness of the decomposition process. This criterion is instrumental in determining the optimal number of modal components \(k\), which is critical for achieving high fidelity in the reconstruction of the original signal. The methodology can be outlined in several systematic steps:
\begin{enumerate}
\item \textbf{Decomposition: }Initially, the original signal \(y\) is decomposed into various modal components using VMD for different potential values of \(k\). This step is essential to explore the range of possible decompositions and to prepare for subsequent analysis.
\item \textbf{Reconstruction: }Following the decomposition, these modal components are reassembled to form the reconstructed signal \(y_0\). This step tests the completeness of the original decomposition by attempting to recreate the initial signal using the modal components derived.
\item \textbf{Evaluate MIC: }The Mutual Information Criterion (MIC) is then calculated to assess the degree of similarity between the original signal \(y\) and the reconstructed signal \(y_0\). The MIC, denoted as \( {\text{MICyy}_0} \), evaluates the mutual information, providing a statistical measure of how well the reconstructed signal represents the original.
\item \textbf{Optimization: }The final step involves selecting the value of \(k\) that yields the maximum MIC value. An MIC value approaching 1 signifies minimal information loss during the reconstruction process, indicating that the decomposition has been optimally performed with that particular number of modal components.
\end{enumerate}

By utilizing the MIC in this manner, VMD can be fine-tuned to ensure minimal information loss and optimal reconstruction quality. This approach not only enhances the analytical capabilities of VMD but also ensures that the signal processing is both precise and efficient, leading to more accurate interpretations and outcomes in practical applications.

\subsubsection{Fuzzy Entropy Approach}

Fuzzy Entropy ($\text{FE}$) serves as a prominent technique tailored for assessing the complexity inherent in temporal series, offering a profound insight into the intricacies of the data structure \cite{71}. A salient feature of $\text{FE}$ is its adaptive nature, which allows for a seamless evolution in accordance with changes in its prescribed parameters. This adaptability confers upon $\text{FE}$ a robust defense mechanism against noise-induced perturbations, ensuring its integrity even in the presence of external disturbances.

For a time series of length \( n \), the $\text{FE}$ algorithm is seamlessly integrated within the framework governed by fuzzy membership functions. This methodological integration results in the subsequent formulation:
\begin{align}
\ln{\text{D}(x)} = -\ln(2)\left(\cfrac{x}{r}\right)^{2}
\end{align}
Here, \( r \) represents the similarity tolerance, and \( x \) is equated to \( d^{m}_{ij} \), the distance between vectors that reconstruct the time series in an \( m \)-dimensional phase space. For \( i, j = 1, 2, \ldots, n - m + 1 \), taking an average over each \( i \) in \( \text{D}^{m}_{ij} \) produces the average similarity function:
\begin{equation}
\phi^m(r) = \cfrac{1}{\text{N}-m+1} \displaystyle\int\limits_{i=1}^{\text{N}-m+1}\left(\cfrac{1}{\text{N}-m} \displaystyle\int\limits_{j=1, j \neq i}^{\text{N}-m+1} \text{D}_{i j}^m\right)
\end{equation}
$$
\downarrow
$$
\begin{equation}
\text{FE}(m, r, n) = \ln\left(\cfrac{\Phi{}^{m}(r)}{\Phi{}^{m+1}(r)}\right)
\end{equation}

\subsection{DS Encoder Informer: Encoder-Based Optimized Informer}
In our research on long-term financial time series prediction, we've established a rolling prediction setting with fixed window sizes, linking input at time \( t \) with its corresponding sequence prediction output. This model supports output lengths that exceed the length of previous inputs and is not limited to univariate feature dimensions. At the core of many popular models, including our proposed DS Encoder Informer Architecture, is the encoder-decoder architecture designed to encode the input representation \( X_t \) into a hidden state \( H_t \), and then decode this state to produce the output \( Y_t \). This involves a process known as "dynamic decoding," where the decoder iteratively infers by calculating a new hidden state \( h_{t(k+1)} \) from the previous state \( h_{tk} \) and the required outputs at step \( k \), and predicts the next sequence \( y_{t(k+1)} \). For an overview of this architecture and more details, please refer to our framework diagram.

\subsubsection{Temporal Feature Embedding Mechanism}
In the context of financial time series forecasting, the integration of hierarchical timestamps (like weeks, months, and years, as well as holidays and significant events) is vital for improving predictive accuracy. Traditional self-attention mechanisms often struggle with effectively processing this rich temporal information, which can degrade the interaction between query and key in encoder-decoder architectures.

\begin{figure}[t!]
\centering
\includegraphics[width=\columnwidth]{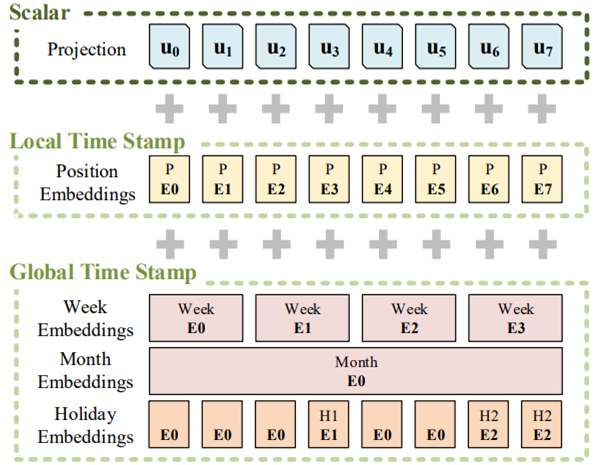}
\xdef\xfigwd{\the\wd\figbox}
\usebox\figbox
\caption{The input embedding for the Informer model consists of three distinct components: scalar projection, local timestamp embeddings (positional), and global timestamp embeddings. }
\vspace{5pt}
\raggedright
\footnotesize{\textbf{Note: }The figure is quoted from the literature \cite{39}}
\label{cite}
\end{figure}

To address these challenges, our study introduces a sophisticated input representation that blends both local and global temporal information. Here's how we've refined the methodology:

\begin{enumerate}

\item \textbf{Local Context Encoding: }Each point in the time series, denoted as $\mathcal{X}^t$, retains its local temporal context through fixed positional embeddings. These embeddings are formulated as follows for any position 'pos' within the sequence:

\begin{equation}
\text{Scale}_{j,x}={(2L_x)^{\cfrac{2j}{d_{\text{model}}}}}
\end{equation}
$$
\downarrow
$$
\begin{equation}
\begin{aligned}
\begin{cases}
\mathrm{PE}_{(\text{pos}, 2j)} = \sin\left[\cfrac{\text{pos}}{\text{Scale}_{j,x}}\right] \\\\
\mathrm{PE}_{(\text{pos}, 2j+1)} = \cos\left[\cfrac{\text{pos}}{\text{Scale}_{j,x}}\right]
\end{cases}
\end{aligned}
\end{equation}
where $j$ is an index from 0 to $\left\lfloor d_{\text{model}} / 2 \right\rfloor$.

\item \textbf{Global Timestamp Embedding: }We utilize a set of learnable embeddings for global timestamps. These embeddings account for a variety of significant temporal markers, with a vocabulary designed to recognize 60 unique timestamps up to minute-level precision $\mathrm{SE}_{(\text{pos})}$ tailored to capture broader temporal dynamics.

\item \textbf{Feature Vector Transformation and Integration: }Each feature $\mathbf{x}_i^t$ at time $t$ and position $i$ is transformed into a higher dimensional space using a one-dimensional convolutional operation, producing $\mathbf{u}_i^t$. The complete input vector for the model is constructed by combining these transformed feature vectors with both local and global embeddings:
\begin{equation}
\mathcal{X}_{\text{feed}[i]}^t = \alpha \mathbf{u}_i^t + \mathrm{PE}_{(i)} + \sum_p \mathrm{SE}_{(i, p)},
\end{equation}
where $\alpha$ is a scaling factor set to 1 to maintain balance between the original features and the embedded representations.
\end{enumerate}
The visualisation of this process is visible mainly in Fig \ref{cite}, This new approach not only ensures effective encoding of both local and global temporal details but also enhances the model's capability to forecast with higher accuracy and computational efficiency.

\subsubsection{Multi-head Distributed Sparse Attention Mechanism (Distributed Informer)}
In this study, which explores new dimensions beyond traditional time series forecasting methods such as wind or power forecasting, financial time series forecasting leverages a dual-feature framework. The primary types of features include economically significant indicators, such as trend indices and oscillation indicators. These are formally represented as follows:
\begin{equation}
\begin{cases}
\mathcal{X}^t = \left\{\mathbf{x}_1^t, \ldots, \mathbf{x}_{L_x}^t \mid \mathbf{x}_i^t \in \mathbb{R}^{d_x}\times 1\right\}
\\\\
\mathcal{Y}^t = \left\{\mathbf{y}_1^t, \ldots, \mathbf{y}_{L_y}^t \mid \mathbf{y}_i^t \in \mathbb{R}^{d_{y}\times 1}\right\}
\end{cases}
\end{equation}
The secondary type involves features extracted directly from the time series through advanced computational techniques such as Fourier transformations. This approach diverges from prior models that typically merged these features without considering their inter-relationships. Our findings highlight that understanding the intricate correlations between these types of indicators is pivotal for enhancing the accuracy of financial forecasts.

During the initial stages of our modeling process, we incorporate three core components: the 'key', 'query', and 'value'. These elements are integrated as follows:
\begin{equation}
[k^{L \times d_k}, q^{L \times d_q}, v^{L \times d_v}]
\end{equation}
where \(L\) represents the sequence length, and \(d_K\), \(d_Q\), and \(d_V\) denote the dimensions of the key, query, and value features, respectively. The alignment and integration with the learnable weight matrices are detailed in:
\begin{equation}
[\omega_{k}^{d_k\times d}, \omega_{a}^{d_q\times d}, \omega_{v}^{d_v\times d}] \rightarrow \begin{cases} 
K^{L \times d}=k w_k \\\\
Q^{L \times d}=q w_a \\\\
V^{L \times d}=v w_v \end{cases}
\end{equation}

This integration fosters a dense and structured data representation, essential for the efficiency of the Distributed Informer, our advanced adaptation of the traditional multi-head attention mechanism. This model excels in processing sparse datasets and enhances predictive performance significantly. The attention dynamics are quantitatively defined as:
\begin{equation}
\mathcal{A}\left(\mathbf{q}_i, \mathbf{K}, \mathbf{V}\right) = \displaystyle\sum\limits_j \cfrac{k\left(\mathbf{q}_i, \mathbf{k}_j\right)}{\displaystyle\sum\limits_l k\left(\mathbf{q}_i, \mathbf{k}_l\right)} \mathbf{v}_j = \mathbb{E}_{p\left(\mathbf{k}_j \mid \mathbf{q}_i\right)}\left[\mathbf{v}_j\right]
\end{equation}
where 
\begin{equation}
p(\mathbf{k}_j \mid \mathbf{q}_i) =\cfrac{ k(\mathbf{q}_i, \mathbf{k}_j) }{ \displaystyle\sum\limits_l k(\mathbf{q}_i, \mathbf{k}_l)}
\end{equation}
utilizes an asymmetric exponential kernel, enhancing the model’s sensitivity to critical features. Furthermore, the computational complexity is also addressed:

\begin{equation}
K L(q \| p) = \ln \displaystyle\sum\limits_{l=1}^{L} \cfrac{e^{^{\mathbf{q}_i \mathbf{k}_l^{\top}}}}{\sqrt{d}} - \cfrac{1}{L} \displaystyle\sum\limits_{j=1}^{L_K} \cfrac{\mathbf{q}_i \mathbf{k}_j^{\top}} {\sqrt{d}} - \ln{L_{K}}
\end{equation}

The self-attention mechanism refines the model's predictions by evaluating output probabilities, thereby enhancing forecasting accuracy through an optimized attention distribution in sparsely populated data environments.

\subsubsection{Stacked Informer Structure}
\begin{figure*}[t!]
  \centering
  \includegraphics[width=\textwidth]{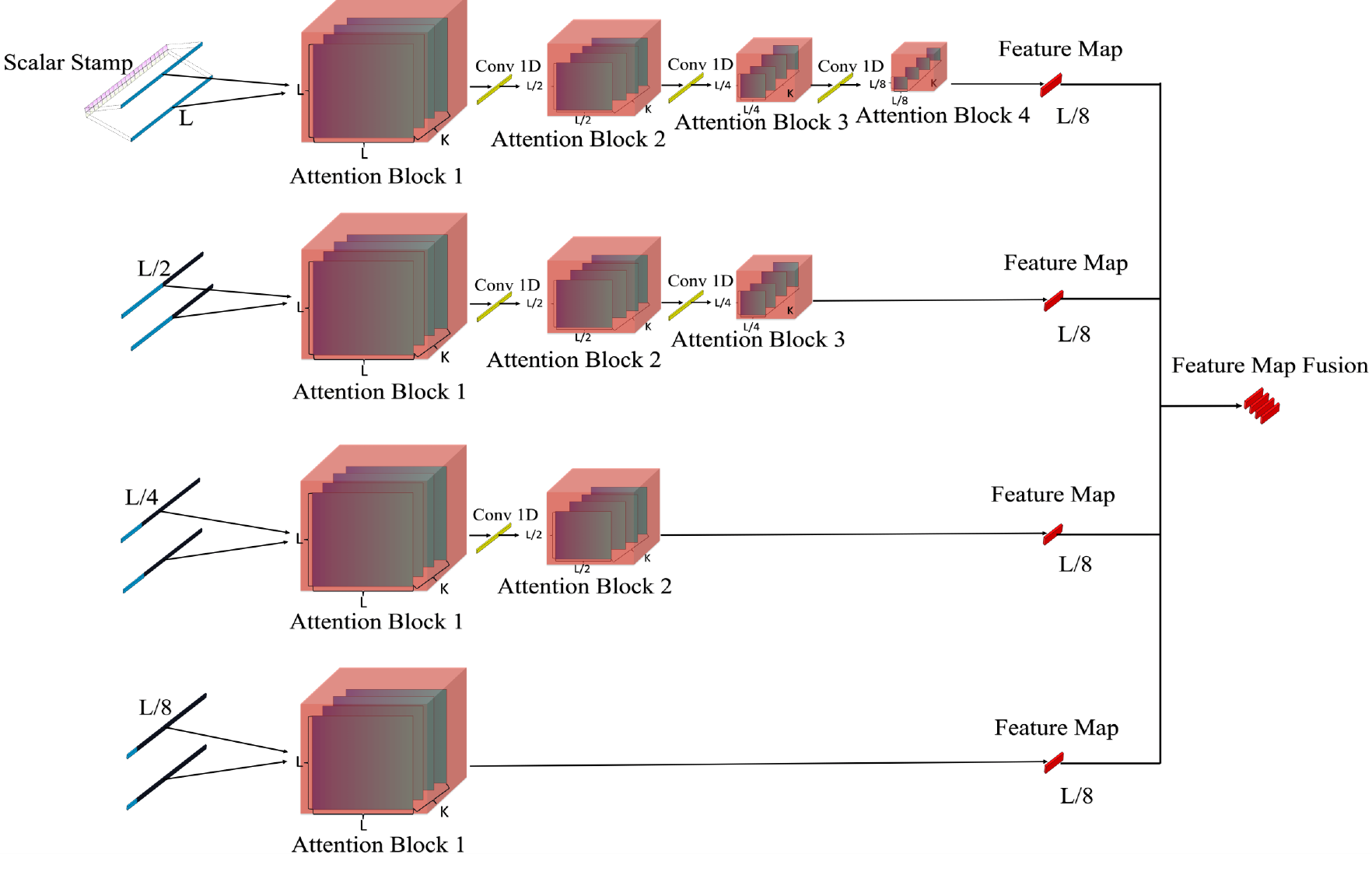}
  \xdef\xfigwd{\the\wd\figbox}
  \usebox\figbox
  \caption{Stacked Informer Structure with 4 Layers.}
  \label{SI}
\end{figure*}

The Stacked Informer architecture, as depicted in Fig \ref{SI}, is an advanced, tiered variant of the multi-layer Informer network specifically engineered for long-term time series forecasting. This structure begins with the original long time-series data, denoted as \( L \). Initially, \( L \) is halved to \( L/2 \), and then quartered to \( L/4 \). The sequence \( L \) is processed through three attention blocks and two convolution layers, while the \( L/2 \) sequence goes through two attention blocks and one convolution layer. The \( L/4 \) sequence is handled by a single attention block. Subsequently, features extracted from all sequences at the \( L/8 \) scale are consolidated into a unified feature map, which is then transmitted to the decoding module.

This Stacked Informer design effectively addresses several critical aspects of long-term forecasting. First, it focuses on efficiently distilling intrinsic temporal features from extensive time-series data. Second, it enhances the robustness and resilience of the forecasts produced. Additionally, the architecture supports parallel processing, thereby improving prediction efficiency. The design utilizes a multi-scale input representation, wherein different temporal scales are processed through distinct pathways. This strategy ensures that the model captures a broad spectrum of features across varying time scales, thereby enriching the informational content essential for long-term forecasting tasks. Moreover, by integrating features from diverse scales into a coherent feature map, the model effectively accounts for interactions and dependencies across these scales.

Each scale within the Stacked Informer employs attention blocks, based on the Transformer architecture, to capture long-distance dependencies within the input sequences. Furthermore, convolutional layers are used to extract local features, which are then integrated with the outputs from the attention blocks to enhance the predictive capabilities of the model.

Overall, the Stacked Informer architecture, with its layered and multi-scale approach, is adept at capturing complex patterns and dependencies within long time series, leading to more precise predictions. This structured design not only reduces the number of attention blocks needed—halving them from four to two for reducing \( L \) to \( 1/32L \)—but also shifts the computation from serial to parallel, optimizing for GPU computations. This enhances computational efficiency, integrates global attention mechanisms with local information processing, and minimizes the introduction of noise, thus reducing model complexity and the risk of overfitting. These enhancements allow the Stacked Informer to deliver superior performance in long-term time series forecasting.

\subsection{Other Methods and Optimisation}
\subsubsection{GC-Enhanced Adam Optimizer}
Optimization strategies are of paramount importance in the effective and efficient training of Deep Neural Networks ($\text{DNNs}$). While traditional optimizers such as Stochastic Gradient Descent ($\text{SGD}$) and Adam have been widely adopted in $\text{DNNs}$, this work incorporates the novel Gradient Centralization ($\text{GC}$) technique into the Adam optimizer, aiming to further enhance its efficacy.

The $\text{GC}$ technique was pioneered by Yong et al. in 2020 \cite{72}. Unlike other methods that modify the learning rates or adaptively adjust weights, $\text{GC}$ directly acts upon the gradients by centralizing them, effectively making their mean values zero. One can draw parallels between $\text{GC}$ and projected gradient descent, though the former operates within the constraints of a specific loss function. This centralization is believed to foster a more stable and efficient training process.

Given a weight vector \( \mathbf{w} \), with corresponding gradients denoted as \( \nabla w_i \text{L} \) (where \( i \) spans from 1 to \( \text{N} \)), the $\text{GC}$ operator, \( \phi_{\text{GC}} \), is formally defined as:

\begin{equation}
\mu \nabla_{w_i} \text{L} = \cfrac{1}{\text{M}} \displaystyle\sum\limits_{j=1}^\text{M} \text{W}_{i, j} \text{L}
\end{equation}
$$
\downarrow
$$
\begin{equation}
\phi_{\text{GC}}\left(\nabla_{w_i} L\right) = \nabla_{w_i} \text{L} - \mu \nabla_{w_i} \text{L}
\end{equation}
Where \( \text{M} \) denotes the dimensionality, and \( \text{L} \) is the objective function. The essence of this formulation is the subtraction of the computed mean of column vectors in the weight matrix from each individual column vector, achieving gradient centralization.

For a more matrix-oriented representation, one can express the aforementioned operation as:
\begin{equation}
\text{P} = \text{I} - e e^\text{T}
\end{equation}
$$
\downarrow
$$
\begin{equation}
\phi_{\text{GC}}\left(\nabla_\text{W} \text{L}\right) = \text{P} \nabla_{\text{W}} \text{L}
\end{equation}
In this formulation, \( \mathbf{W} \) is the weight matrix, \( \mathbf{P} \) is the projection matrix that lies within the hyperplane defined by a normal vector in the weight space, and \( \mathbf{P} \nabla \mathbf{W} \text{L} \) is the gradient projected onto said hyperplane. Once the centralized gradient \( \phi_{\text{GC}}(\nabla \mathbf{\text{W}}\text{L}) \) is obtained, it can be seamlessly used for weight matrix updates.

One potential enhancement to consider when employing $\text{GC}$ with Adam or other adaptive optimizers is the careful tuning of hyperparameters. While $\text{GC}$ aims to stabilize the training process, the adaptive nature of optimizers like Adam means that they adjust learning rates based on past gradient moments. Combining these two methodologies might necessitate a recalibration of hyperparameters to achieve optimal results.

\subsubsection{Dynamic Loss Function}

The adaptive loss function, a novel approach to loss function formulation, was pioneered in the work of Barron \cite{73}. This approach integrates a continuous parameter signifying robustness into the traditional loss function. Throughout the optimization phase of model training, this dynamic loss function refines the robustness parameters in tandem with the loss minimization procedure, enhancing prediction precision. The generalized loss function is mathematically described as:
\begin{equation}
f(z, \beta, c) = \cfrac{|\beta-2|}{\beta}\left[\left(\cfrac{{z}^2}{|\beta-2|c^2}+1\right)^{\cfrac{\beta}{2}}-1\right]
\end{equation}

In this formulation, the variable \(z\) quantifies the residual between observed and predicted values. The positive scalar \(c > 0\) adjusts the curvature of the quadratic component at \(x = 0\), while \(\beta\) serves as the adaptable parameter steering robustness.

A detailed examination of the equation reveals that the behavior of the adaptive loss function is intricately tied to variations in \(\beta\). Different settings of \(\beta\) manifest in the loss function as in the Table \ref{loss}

\begin{table}[ht]
\centering
\caption{Loss Function Value with $ \aleph = \cfrac{1}{2} \left(\cfrac{z}{c}\right)^2$}
\label{loss}
\renewcommand{\arraystretch}{2} 
\begin{tabular}{|c|c|}
\hline
\textbf{$f(z, \beta, c)$ Value} & \textbf{Condition} \\
\hline
$\aleph{} $ & $\beta=2$ \\
$\log \left(\aleph{}  + 1\right)$ & $\beta=0$ \\
$\sqrt{2\aleph{} + 1} - 1$ & $\beta=1$ \\
$1-e^{-\aleph{} }$ & $\beta=-\infty$ \\
$\cfrac{|\beta-2|}{\beta}\left[\left(\cfrac{2\aleph}{|\beta-2|} + 1\right)^{\cfrac{\beta}{2}} - 1\right]$ & \text{Otherwise} \\
\hline
\end{tabular}
\end{table}

It is noteworthy that this adaptive loss function is versatile, enveloping a spectrum of loss functions such as Mean Squared Error (MSE), Cauchy, Charbonnier, and Welsch. This versatility is attributed to the modulation of the \(\beta\) parameter, which grants the flexibility to customize the loss function per specific modeling needs.

Incorporating such a loss function into long-term time series forecasting models can be advantageous. Its adaptability allows for nuanced error handling, potentially leading to improved model robustness and sensitivity to various data patterns, especially in scenarios with non-Gaussian noise or outliers.

\subsubsection{ Evaluation Indicators}
The mean absolute error (MAE), mean square error (MSE), root mean square error (RMSE), and coefficient of determination ($\text{R}^2$) are used as indicators for the prediction performance of our Enhanced LFTSformer and all the other models. The formulas of each indicators are as follow in the Table \ref{indicators}: 

\begin{table}[ht]
\centering
\caption{ Evaluation indicators}
\label{indicators}
\renewcommand{\arraystretch}{3} 
\begin{tabular}{|c|c|}
\hline
\textbf{Indicator} & \textbf{Formula} \\
\hline
MAE & $\cfrac{1}{N} \displaystyle\sum\limits_{t=1}^N \left| q_{\text{true}}(t) - q_{\text{pred}}(t) \right|$ \\

MSE & ${\cfrac{1}{N} \displaystyle\sum\limits_{t=1}^N\left(q_{\text {true }}(t)-q_{\text {pred }}(t)\right)^2}$ \\

RMSE & $\sqrt{\cfrac{1}{N} \displaystyle\sum\limits_{t=1}^N\left(q_{\text {true }}(t)-q_{\text {pred }}(t)\right)^2} $ \\

$\text{R}^2$ & $1-\cfrac{\displaystyle\sum\limits_{t=1}^N\left(q_{\text {true }}(t)-q_{\text {pred }}(t)\right)^2}{\displaystyle\sum\limits_{t=1}^N\left(q_{\text {true }}(t)-\bar{q}\right)^2}$ \\

\hline
\end{tabular}
\end{table}

In addition, in order to be able to analyse and compare the advantages and disadvantages of the models more intuitively, we will draw a total of three heat maps for MSE, RMSE, and $\text{R}^2$ respectively. When MSE, RMSE is smaller or $\text{R}^2$ is larger, the colour of the heat map will be darker. We will also blend the three heatmaps by colour, and the darker the colour of the resulting blended heatmap means that the corresponding model/stock forecast is better.

\section{Experiment Process and Results} \label{4}
\subsection{Experiment Conditions \& Configuration}
\subsubsection{Experiment Environment}
Regarding the experimental configuration, our comprehensive investigations are carried out utilizing a computer system endowed with dual NVIDIA GeForce RTX 4090 GPUs, accompanied by an Intel13th Gen Intel(R)Core(TM)i9-13900K CPU and a robust 128GB RAM. 

The operational environment is Wiundows 11 23H2 operating system, while the deep learning framework employed throughout our endeavors remains to be PyTorch.

\subsubsection{Experiment Data}
For the data used in the experiment, we have a few points to note
\begin{enumerate}
\item \textbf{Data Selection: }Our ablation experiments as well as the comparison experiments are conducted on the data of Pudong Development Bank stock (SHA: 600000). After completing the training of the model, we will conduct stability experiments on the remaining 21 stocks.
\item \textbf{Data Segmentation: }For all stock data, we set the first 90\% of the data as the training set and the last 10\% as the test set. As shown in Fig \ref{data}.
\begin{figure*}[t!]
  \centering
  \includegraphics[width=\textwidth]{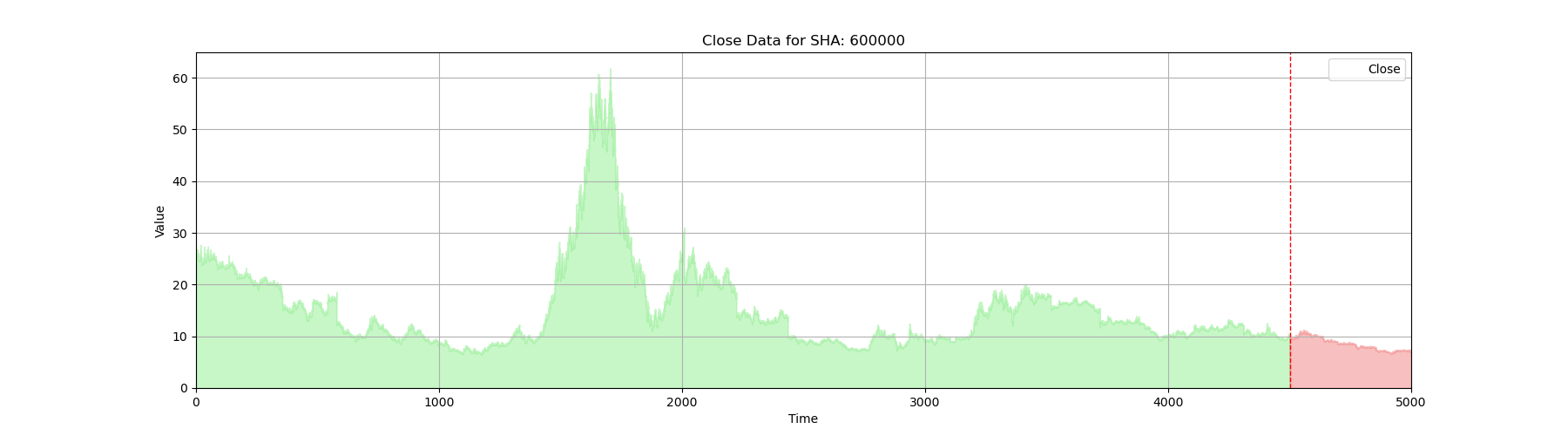}
  \xdef\xfigwd{\the\wd\figbox}
  \usebox\figbox
  \caption{Close Data of SHA: 600000, Split Training and Test Sets at 90\% Ratio}
  \label{data}
\end{figure*}
\item \textbf{Parameters of prediction: }For all stocks and different models, we are predicting the Close parameter, while the other parameters are Indexing and Manipulation as described in Section \ref{1}.
\item \textbf{Time Scaling: }Since there are some differences in the time span of different stocks, for the sake of convenience, we set the start time as 0 time unit, the end time as 5000 time unit, and the other data are scaled to the corresponding time unit.
\end{enumerate}

\subsubsection{Training Parameter Configuration}
After rigorous experimentation and meticulous debugging, we have judiciously chosen a set of parameters that optimally enhances the predictive capabilities of our model. The salient parameters integral to our model's architecture are delineated in Table \ref{tab_parameters}.

\begin{table}[h]
\caption{Model Parameters Configuration}\label{tab_parameters}%
\centering
\renewcommand{\arraystretch}{1.3}
\begin{tabular}{@{}cc@{}}
\toprule
\textbf{Parameter} & \textbf{Value} \\
\midrule
Input sequence length & 256 \\
Prediction sequence length & 64-128 \\
Num of encoder layers & 8 \\
Num of decoder layers & 10 \\
Input size of encoder & 5-1 \\
Input size of decoder & 5-1 \\
Num of heads & 2 \\
Dimension of model & 512 \\
Probsparse attention factor & 3.8 \\
Early stopping patience & 10 \\
Dropout & 0.2 \\
Epochs & 200 \\
\hline
\end{tabular}
\end{table}

\subsection{Process of Feature Engineering}
\subsubsection{Determination of K Value and Decomposition of Features}

\begin{figure*}[t!]
\centering
\includegraphics[width=0.85\textwidth]{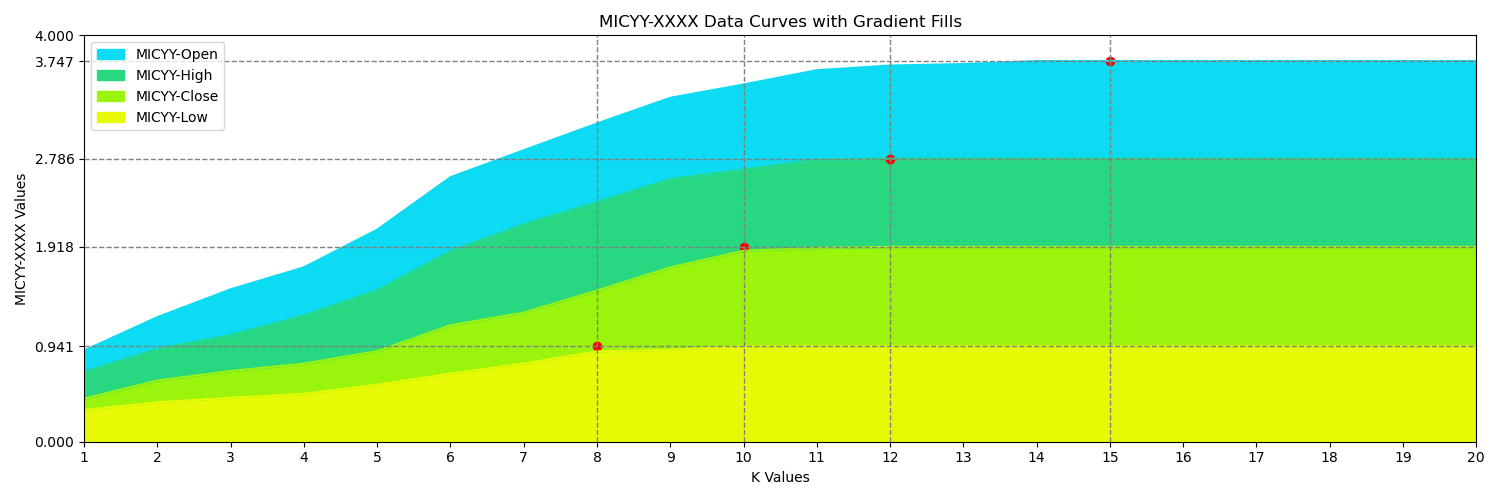}
\xdef\xfigwd{\the\wd\figbox}
\usebox\figbox
\caption{Relationship Between Different Ks and Their Corresponding MICyy.}
\label{MICYY}
\end{figure*}

The \textit{VMD-MIC} method enhances the conventional \textit{VMD} approach by utilizing the computation of the \textit{MICyy0} metric. When the original stock dataset of Open/Close/High/Low is input into the \textit{VMD-MIC} framework, it is decomposed into \( K \) intrinsic mode functions (IMFs). Figure \ref{MICYY} illustrates the relationship between various \( K \) values and their corresponding \textit{MICyy0} results.We can determine the optimum k when the histogram reaches a steady state.

Notably, the \textit{MICyy0} metric remains relatively stable for \( K = 8, 10, 12 \) and \( 15 \). Based on this observation, we make the following selection of $K$ values as shown in the Table\ref{tab_3}:

\begin{table}[h]
\centering
\renewcommand{\arraystretch}{1.3}
\caption{Selection of K Values}\label{tab_3}%
\begin{tabular}{@{}ccc@{}}
\toprule
Indicator & Selection of K for Corresponding Indicator  \\
\midrule
Open      & 15    \\
High    & 10    \\
Close        & 12    \\
Low      & 8    \\
\hline
\end{tabular}
\end{table}

\begin{figure*}[t!]
  \centering
  \includegraphics[width=0.85\textwidth]{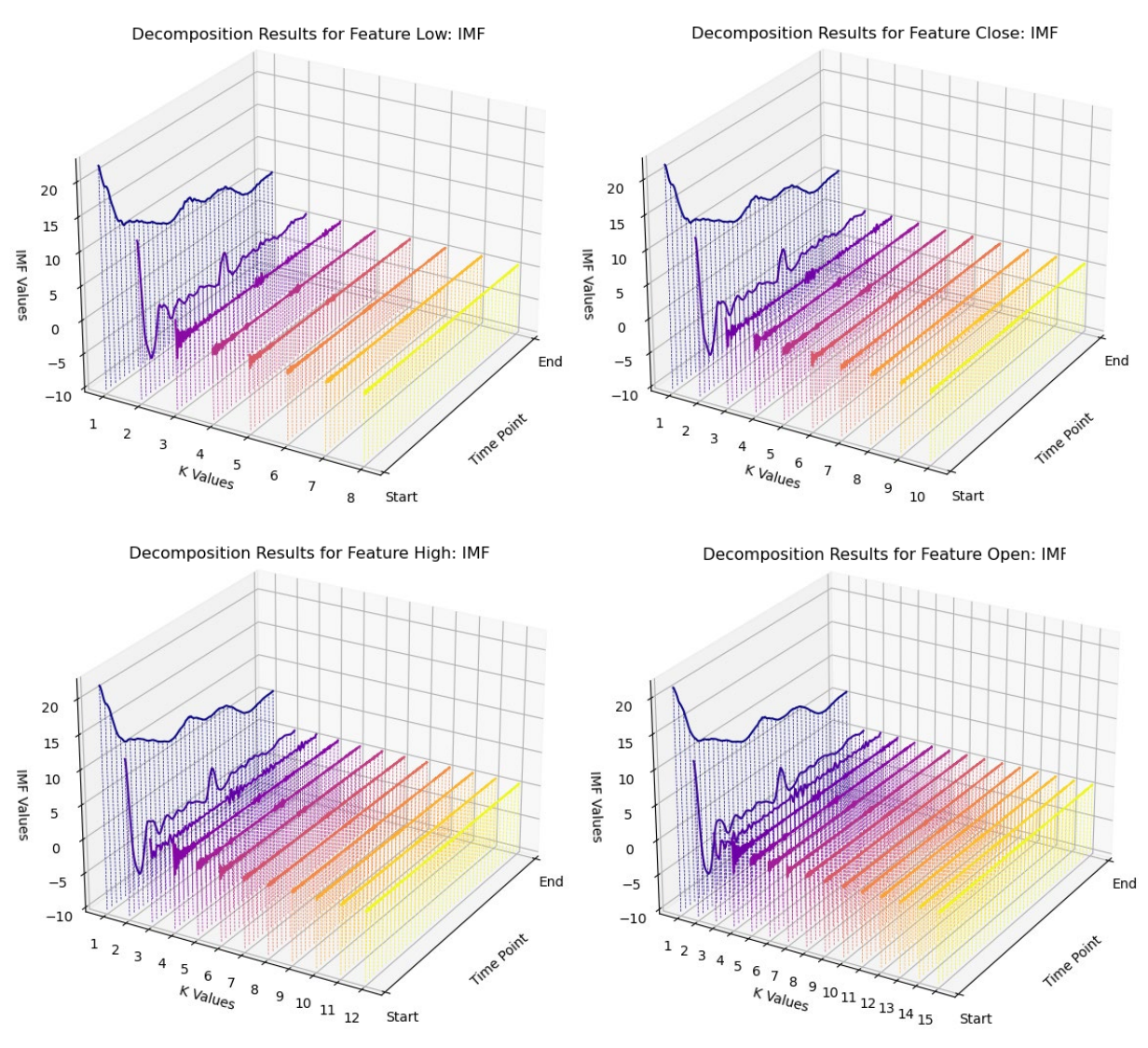}
  \xdef\xfigwd{\the\wd\figbox}
  \usebox\figbox
  \caption{The Decomposition Results For Each Feature: IMFs Values for Each K.}
  \label{IMF}
\end{figure*}

\begin{figure*}[t!]
  \centering
  \includegraphics[width=0.85\textwidth]{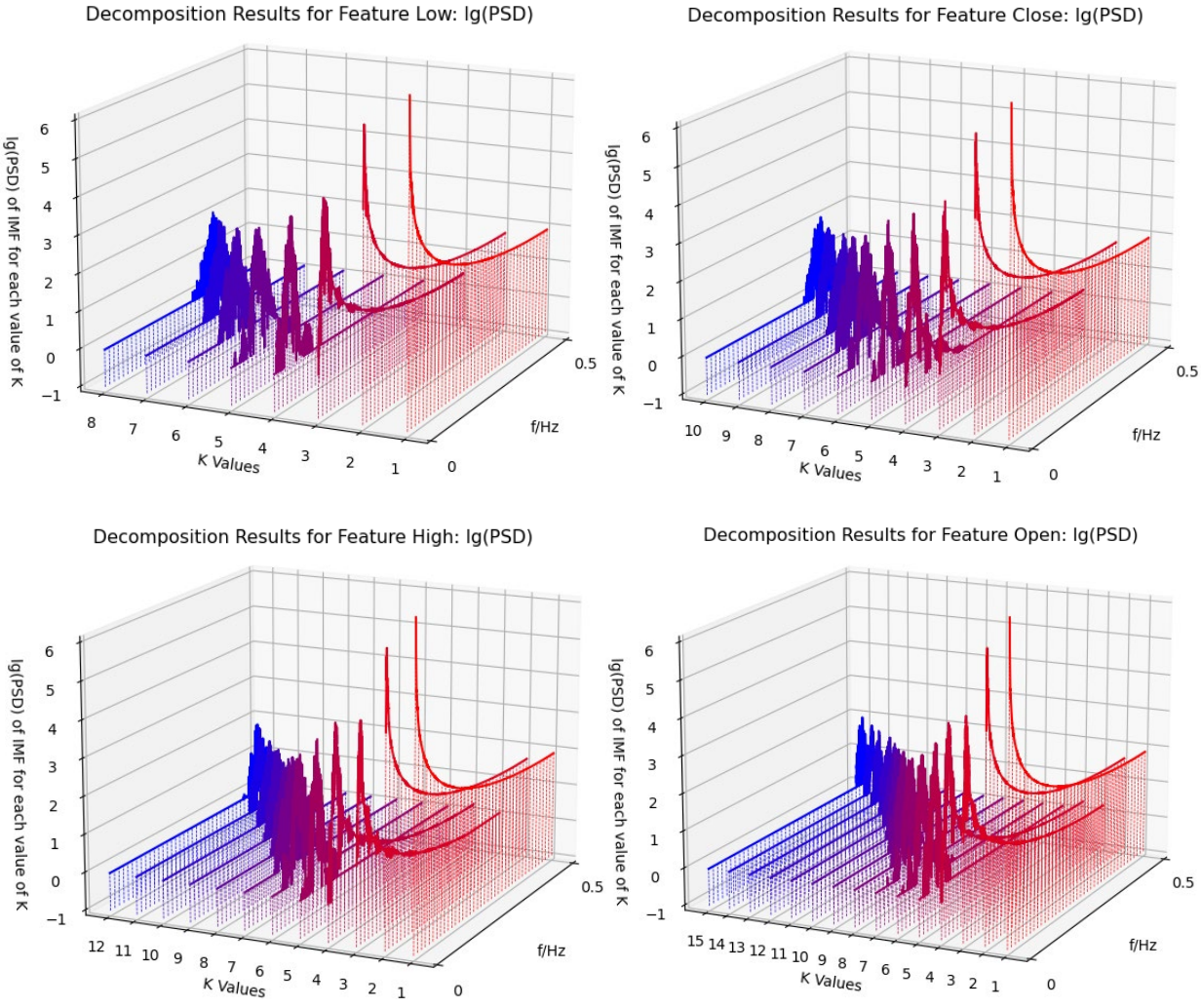}
  \xdef\xfigwd{\the\wd\figbox}
  \usebox\figbox
  \caption{The Dcomposition Results For Each Feature: Power Spectral Density (PSD)
for Each Corresponding IMFs.}
  \label{PSD}
\end{figure*}

Based on this premise, we have delineated the IMF values for the four features: close, high, low, and open prices, post-decomposition at varying \( k \) values as depicted in Fig\ref{IMF}. Correspondingly, we also present their associated Power Spectral Density (PSD) values in Fig\ref{PSD}. It is discerned that with the incremental growth of \( k \) values, their respective trends largely converge. By conducting an in-depth evaluation of the dominant frequencies in these IMF curves, we can discern the efficiency of the VMD-MIC technique in ensuring precise IMF separation.

\subsubsection{Dimensionality Reduction of Features}

\begin{figure*}[t!]
  \centering
  \includegraphics[width=\textwidth]{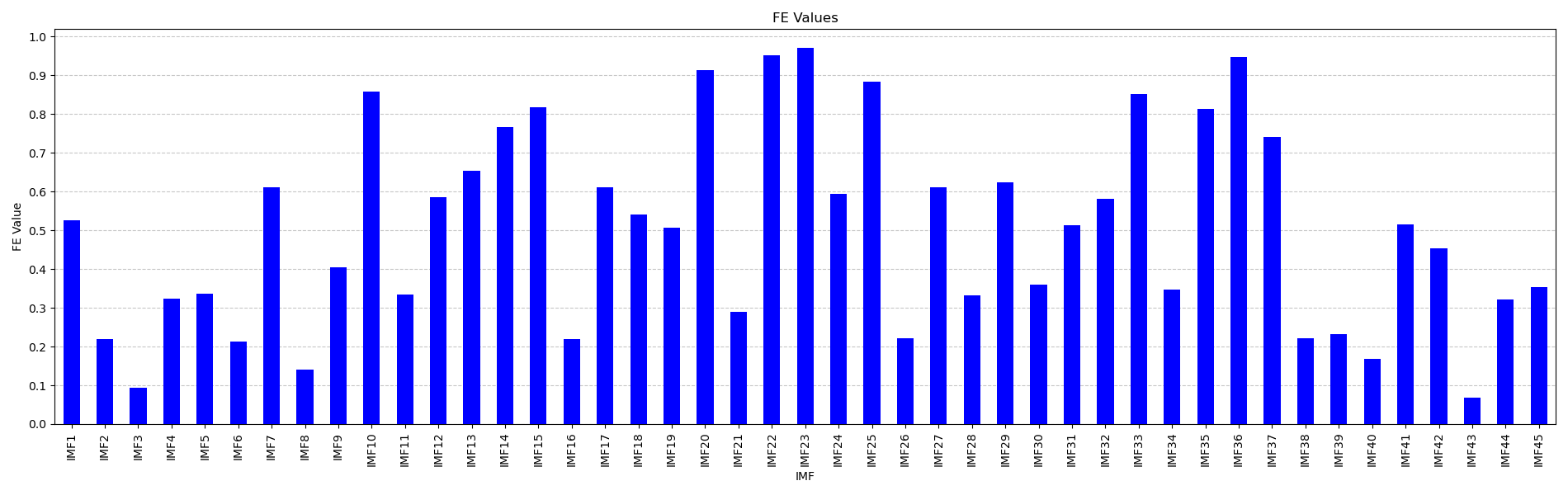}
  \xdef\xfigwd{\the\wd\figbox}
  \usebox\figbox
  \caption{FE Values for 45 IMFs.}
  \label{FE}
\end{figure*}

The original stock dataset was decomposed into intrinsic mode functions (IMFs) with allocations of 10 for close, 12 for high, 8 for low, and 15 for open values. Direct integration of these sub-sequences into the predictive model could exacerbate computational overhead. To address this, we employed Feature Extraction (FE) to quantitatively assess the intricacy of the IMFs, subsequently reconstituting them based on their complexity profiles. Through rigorous research, parameters $m = 3$ and $r = 0.3\text{std}$ were delineated as optimal, striking a balance between model accuracy and computational efficiency. The corresponding visualization in Fig \ref{FE} elucidates the FE values for each IMF.

Upon categorization of IMFs by FE value intervals (e.g., 0.0-0.1, 0.1-0.2, etc.), we crafted new composite elements founded on these FE metrics as shown in the Table \ref{tab_4}.

\begin{table}[h]
\renewcommand{\arraystretch}{1.3}
\caption{New Composite Elements Founded FE Metrics}\label{tab_4}
\centering
\begin{tabular}{@{}ccc@{}}
\toprule
New Feature Grouping & FE Value Range & Selection of $\text{IMF}_{\text{N}}$ (value of N)  \\
\midrule
New Feature 1     & [0.0,0.1]   & 3,43 \\
New Feature 2     & (0.1,0.2]   & 2,8,40 \\
New Feature 3     & (0.2,0.3]   & 6,16,21,26,38,39 \\
New Feature 4     & (0.3,0.4]   & 4,5,11,28,30,34,44,45 \\
New Feature 5     & (0.4,0.5]   & 9,42 \\
New Feature 6     & (0.5,0.6]   & 1,7,12,18,19,24,31,32,41 \\
New Feature 7     & (0.6,0.7]   & 13,17,27,29 \\
New Feature 8     & (0.7,0.8]   & 10,14,37 \\
New Feature 9     & (0.8,0.9]   & 15,25,33,35 \\
New Feature 10    & (0.9,1.0]   & 20,22,23,36 \\
\bottomrule
\end{tabular}
\footnotesize
\begin{flushleft}
Note: New Feature N = $\cfrac{\displaystyle\sum_{\text{Selected N}}\text{IMF}_{\text{N}}}{\text{V}}$, (V=Number of IMFs selected). \\
\end{flushleft}
\end{table}

\subsubsection{Correlation-Based Feature Extraction and Selection}\label{subsubsec3}

\begin{figure*}[t!]
  \centering
  \includegraphics[width=\textwidth]{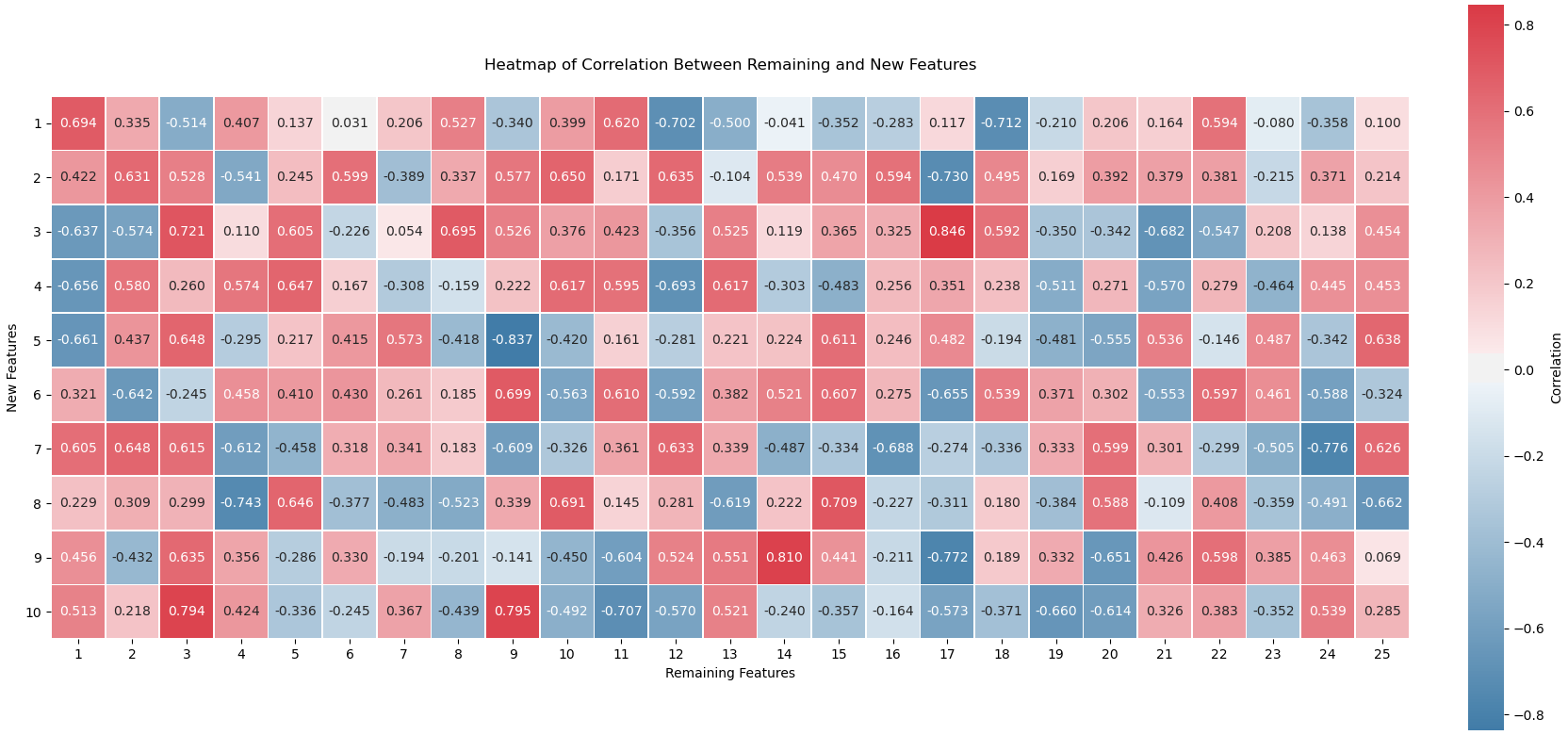}
  \xdef\xfigwd{\the\wd\figbox}
  \usebox\figbox
  \caption{Heatmap For Features.}
  \label{Heatmap}
\end{figure*}

In our endeavor to efficiently extract the residual 25 indicator features, we discerned that the computational efficacy and generalizability of the model could be enhanced by excluding certain irrelevant or redundant attributes from the primary dataset. Consequently, we adopted the Maximal Information Coefficient (MIC) methodology to scrutinize the interrelations between the remaining 25 stock features and each component, extracting quintessential characteristics of each element through their MIC values. The confusion matrix for MIC is depicted in the subsequent Heatmap of Features (Fig \ref{Heatmap}).

From the ensuing visualization, it's palpable that each element manifests distinct influential characteristics, encapsulating both holistic correlations and nuanced local attributes. In pursuit of curating features that bear the utmost correlation with the input variables, we instituted an MIC threshold of 0.5 for each component. The outcomes of this selective input feature determination are delineated in the Table \ref{tab_5} below.

\begin{table}[h]
\centering
\renewcommand{\arraystretch}{1.3}
\caption{Reconstructed Feature Values Based On Heatmap}\label{tab_5}
\begin{tabular}{@{}cc@{}}
\toprule
\textbf{Reconstruct Features} & \textbf{Included Remaining Feature M (M value)} \\
\midrule
Reconstruct Feature 1  & 1, 8, 11, 22 \\
Reconstruct Feature 2  & 2, 3, 6, 9, 10, 12, 14, 16 \\
Reconstruct Feature 3  & 3, 5, 8, 9, 13, 17, 18 \\
Reconstruct Feature 4  & 2, 4, 5, 10, 11, 13 \\
Reconstruct Feature 5  & 3, 7, 15, 21, 25 \\
Reconstruct Feature 6  & 9, 11, 14, 15, 18, 22 \\
Reconstruct Feature 7  & 1, 2, 3, 12, 20, 25 \\
Reconstruct Feature 8  & 5, 10, 15, 20 \\
Reconstruct Feature 9  & 3, 12, 13, 14, 22 \\
Reconstruct Feature 10 & 1, 3, 9, 13, 24 \\
\bottomrule
\end{tabular}
\footnotesize
\begin{flushleft}
Note 1: The features corresponding to all areas greater than 0.5 in the heatmap are selected into the corresponding group.\\
Note 2: $\text{RCF}_{\text{N}} = \text{NF}_{\text{N}}*\left(\displaystyle\prod\limits_{\text{Included M}} \text{RMF}_{\text{M}}*\text{C}_{\text{NM}}\right)$\\
\begin{minipage}{\linewidth}
\vspace{1em}
\centering
\begin{tabular}{cc}
\toprule
\renewcommand{\arraystretch}{1.3}
\textbf{Abbreviation} & \textbf{Meaning} \\
\midrule
$\text{RCF}_{\text{N}}$ & Reconstruct Feature N \\
$\text{NF}_{\text{N}}$  & New Feature N \\
$\text{RMF}_{\text{M}}$ & Remaining Feature M \\
 $\text{C}_{\text{NM}}$ & Correlation Between $\text{NF}_{\text{N}}$ and $\text{RMF}_{\text{M}}$\\
\bottomrule
\end{tabular}
\end{minipage}
\end{flushleft}
\end{table}

\subsection{Results}
In this section, in order to assess the superiority of each module in the model and its overall performance, we conduct and present ablation experiments and comparison experiments on SHA: 600000 stocks and their results. In addition, to assess the generality and robustness of the model, we also conduct stability tests on all other 21 stocks to test our model.

\subsubsection{Ablation Experiment}
The purpose of the ablation study is to assess whether complex hybrid models can enhance prediction accuracy over simpler composite and standalone models. We selected benchmark models including the GC-Enhanced Adam+DS Encoder Informer (also referred to as the Enhanced LFTSformer), the Adam+DS Encoder Informer, and the DS Encoder Informer. Parameter optimization was conducted using a grid search method, where the robustness parameter $\beta$ is adaptively adjusted by the Adam Optimiser. The specific parameters are detailed in Table \ref{tab_parameters}. The sizes of the encoder and decoder inputs are aligned with the number of input variables for the model.

As illustrated in Figure \ref{loss}, the training results for the four models are presented. It is evident that the GC-Enhanced Adam+DS Encoder Informer achieves stability more quickly and exhibits the lowest error rates in both training and testing phases, aligning with our initial expectations.
\begin{figure*}[t!]
  \centering
  \includegraphics[width=0.8\textwidth]{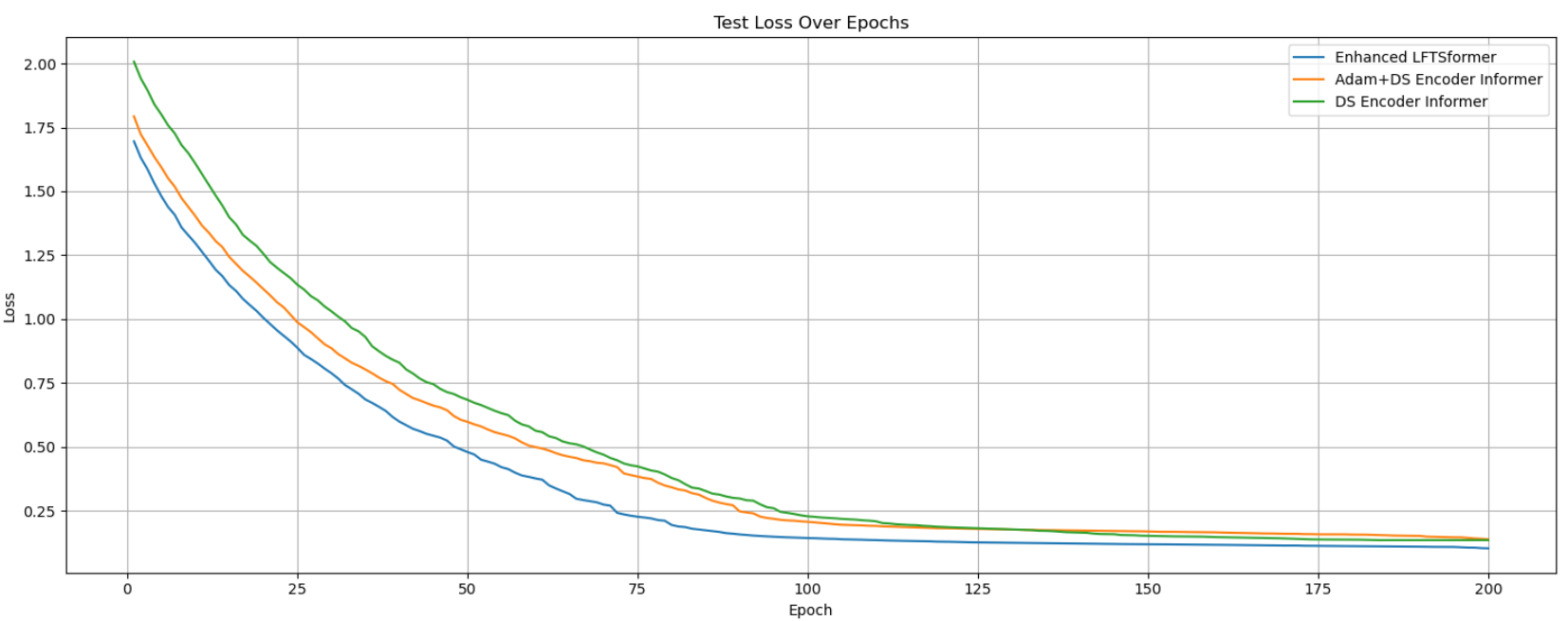}
  \xdef\xfigwd{\the\wd\figbox}
  \usebox\figbox
  \caption{Traning Results For Each Model in Ablation Experiments.}
  \label{loss}
\end{figure*}

\begin{figure*}[t!]
  \centering
  \includegraphics[width=\textwidth]{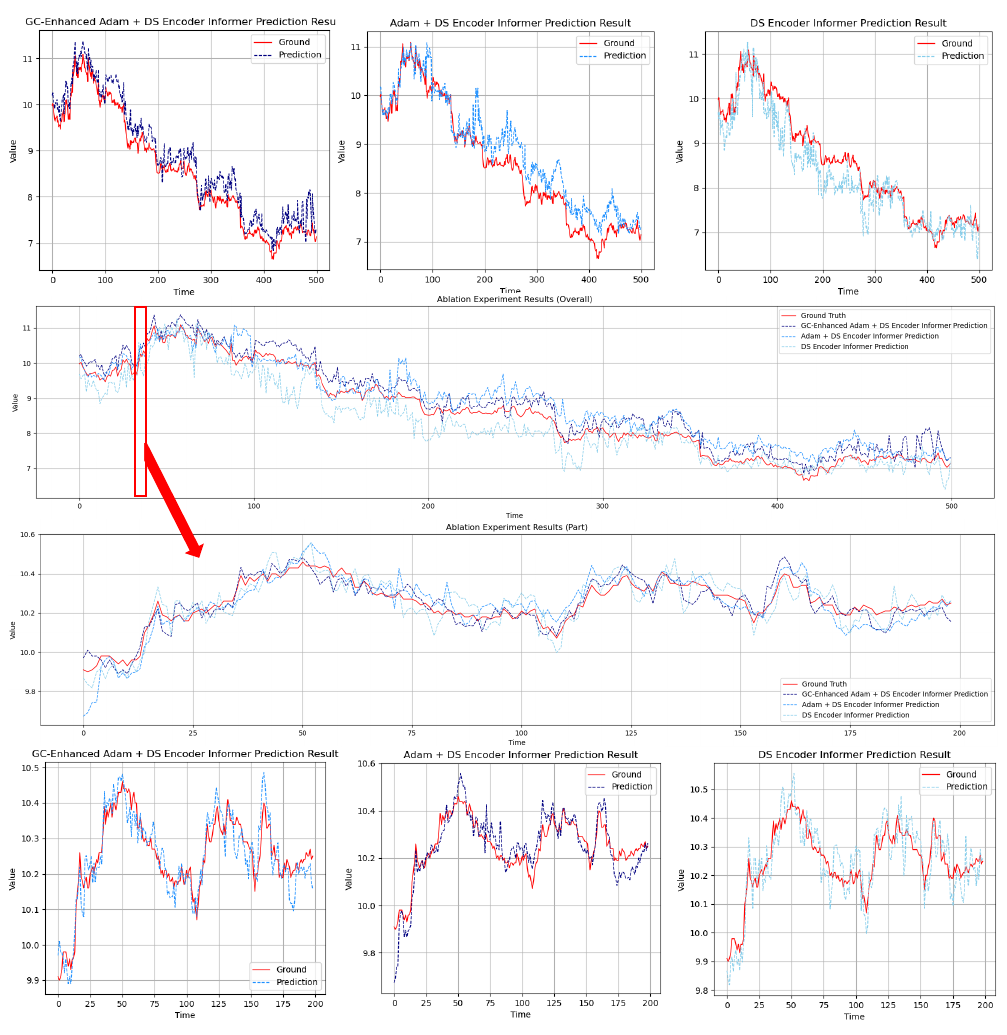}
  \xdef\xfigwd{\the\wd\figbox}
  \usebox\figbox
  \caption{Results of Ablation Experiments for the Whole Test Set and Partial Test Set}
  \label{Disso}
\end{figure*}

Figure \ref{Disso} depicts the results of our ablation experiment. In the top half, we present the predicted outcomes and trends over 500 time units for the bottom 10\% of the dataset. To effectively compare the performance between models and to showcase the performance of individual models, we include integrated prediction plots for all models as well as separate prediction plots for each model.
For the micro-scale analysis, as illustrated in the bottom half of Figure \ref{Disso}, we have selected a smaller time scale (highlighted by the red box in the figure). This particular segment was chosen because the prediction ranges of the models are most closely aligned during this period, covering a total range of 30 minutes × 200 data points for analysis.

\begin{figure*}[t!]
  \centering
  \includegraphics[width=0.87\textwidth]{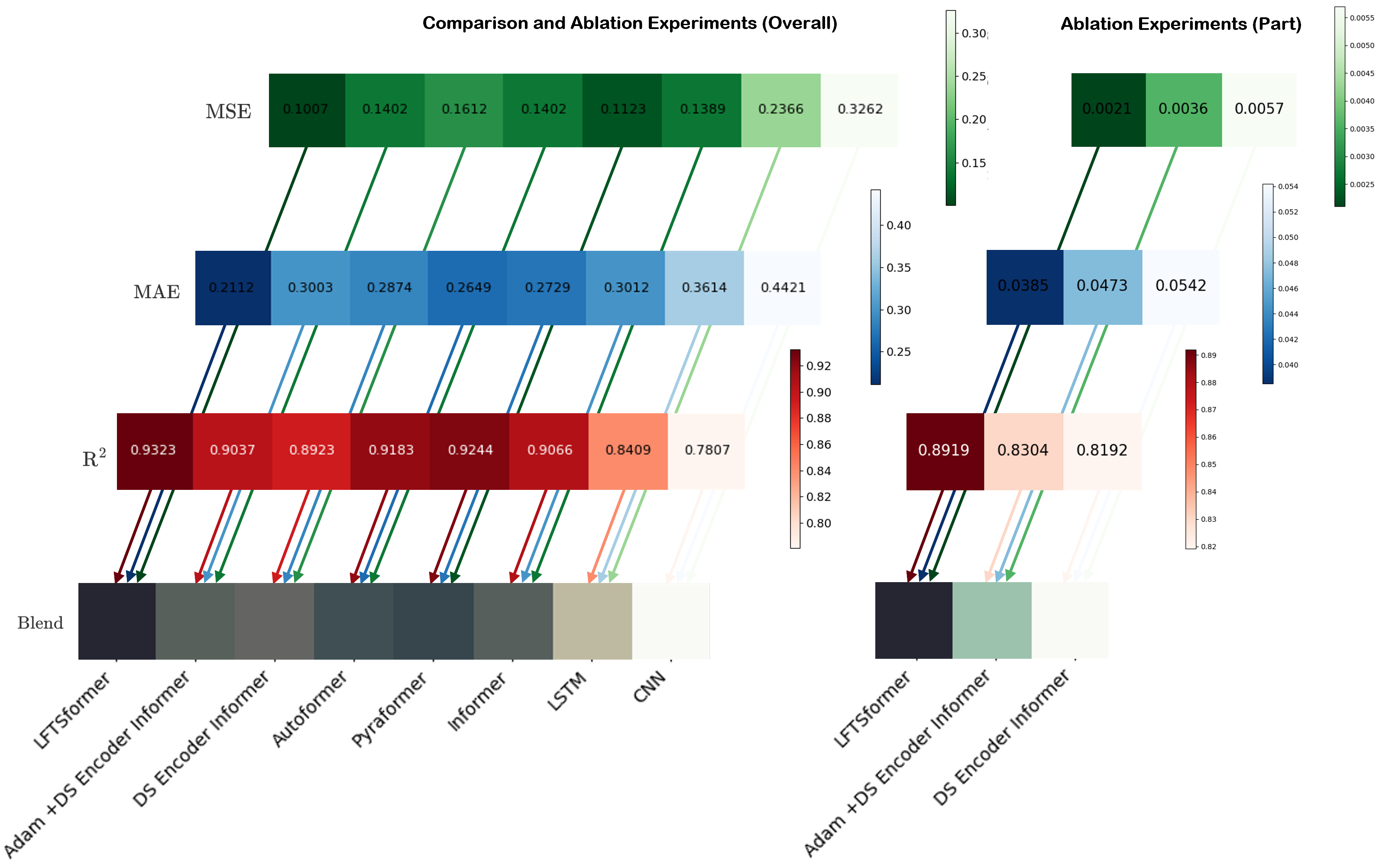}
  \xdef\xfigwd{\the\wd\figbox}
  \usebox\figbox
  \caption{Heat Map for MAE/MSE, $\text{R}^{2}$ and Their Mix for Different Models in Ablation Experiments and Comparison Experiments.  (Darker colours in the blend heat map represent a better combination of predicted performance)}
  \label{XXX}
\end{figure*}

\begin{figure*}[t!]
  \centering
  \includegraphics[width=0.8\textwidth]{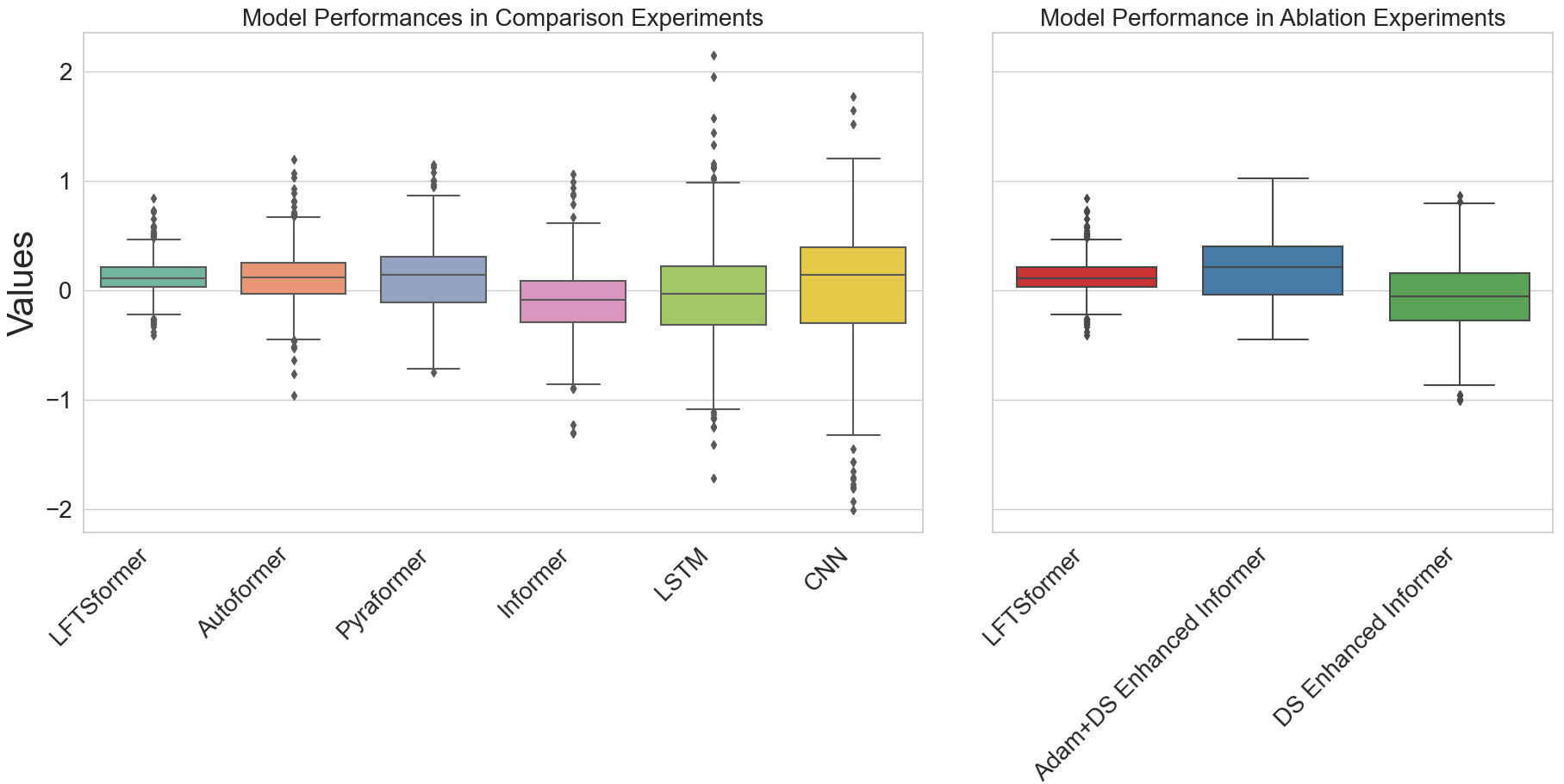}
  \xdef\xfigwd{\the\wd\figbox}
  \usebox\figbox
  \caption{Box Plot of $\text{q}_{real}-\text{q}_{prediction}$ for Different Models in Ablation Experiments and Comparison Experiments}
  \label{Box}
\end{figure*}

In terms of error quantification, we have employed a paradigm incorporating both heatmaps and blended heatmaps to visualize the values of MAE/MSE and $\text{R}^{2}$, as depicted on the right side of Fig\ref{XXX}. Furthermore, to articulate the absolute error, Box Plots are utilized to delineate the distribution and range of the residuals—predicted values subtracted from actual values—and their outliers, as demonstrated on the right side of Fig\ref{Box}. Synthesizing the above, the following analysis can be articulated:

\begin{enumerate}
  \item \textbf{Comprehensive Assessment of Model Performance}: Empirical evidence suggests that the Enhanced LFTSformer model attains a predictive accuracy exceeding 90\% on a holistic scale. Notwithstanding a discernible decrement in accuracy during the transition from macroscopic to microscopic scale analysis, the model's utility and contribution to the field cannot be overstated.
  \item \textbf{Merits of the Adaptive Function in Prediction}: The volatile nature of financial markets, often accentuated by inflection points, poses significant predictive challenges. The incorporation of GC-Enhanced methodology has demonstrably augmented the model's precision. Furthermore, it substantiates the efficacy of adaptive functions in mitigating the impact of market volatilities, thereby reducing predictive errors and uncertainties.
  \item \textbf{Efficiency of Training and the Role of Loss Function}: The pivotal role of a tailored loss function in optimizing both the efficiency and the veracity of model training is incontrovertible. The deployment of our novel adaptive loss function has appreciably curtailed the temporal expenditure required for training the Enhanced LFTSformer model, simultaneously fortifying its robustness and predictive verisimilitude.
\end{enumerate}
Upon contrasting various models, it becomes evident that the Enhanced LFTSformer model possesses the capability to decompose raw stock data into finer granules, thereby delving deeper into the intrinsic dynamics and features of stocks. This intricate exploration furnishes financial decision-makers with predictions that are both precise and timely.


\subsubsection{Comparison Experiment}
To rigorously assess the overall superiority of the proposed model, comparative experiments were conducted. The selection of benchmark models includes not only the classical CNN and LSTM and the foundational Informer but also other state-of-the-art long temporal series forecasting models based on modifications of the Informer/Transformer, such as Autoformer\cite{74} and Pyraformer\cite{75}. Training was carried out on the SHA: 600000 dataset, with all models, including the Enhanced LFTSformer, sharing an identical set of training parameters. The forecast outcomes are presented in Fig\ref{Comp}.

For the comparative analysis of errors across different models, the same methodologies were employed—heatmaps, blended heatmaps, and Boxplots—as shown on the left side of Figs\ref{XXX} and \ref{Box}. From this, we can draw the following conclusions:
\begin{figure*}[t!]
  \centering
  \includegraphics[width=\textwidth]{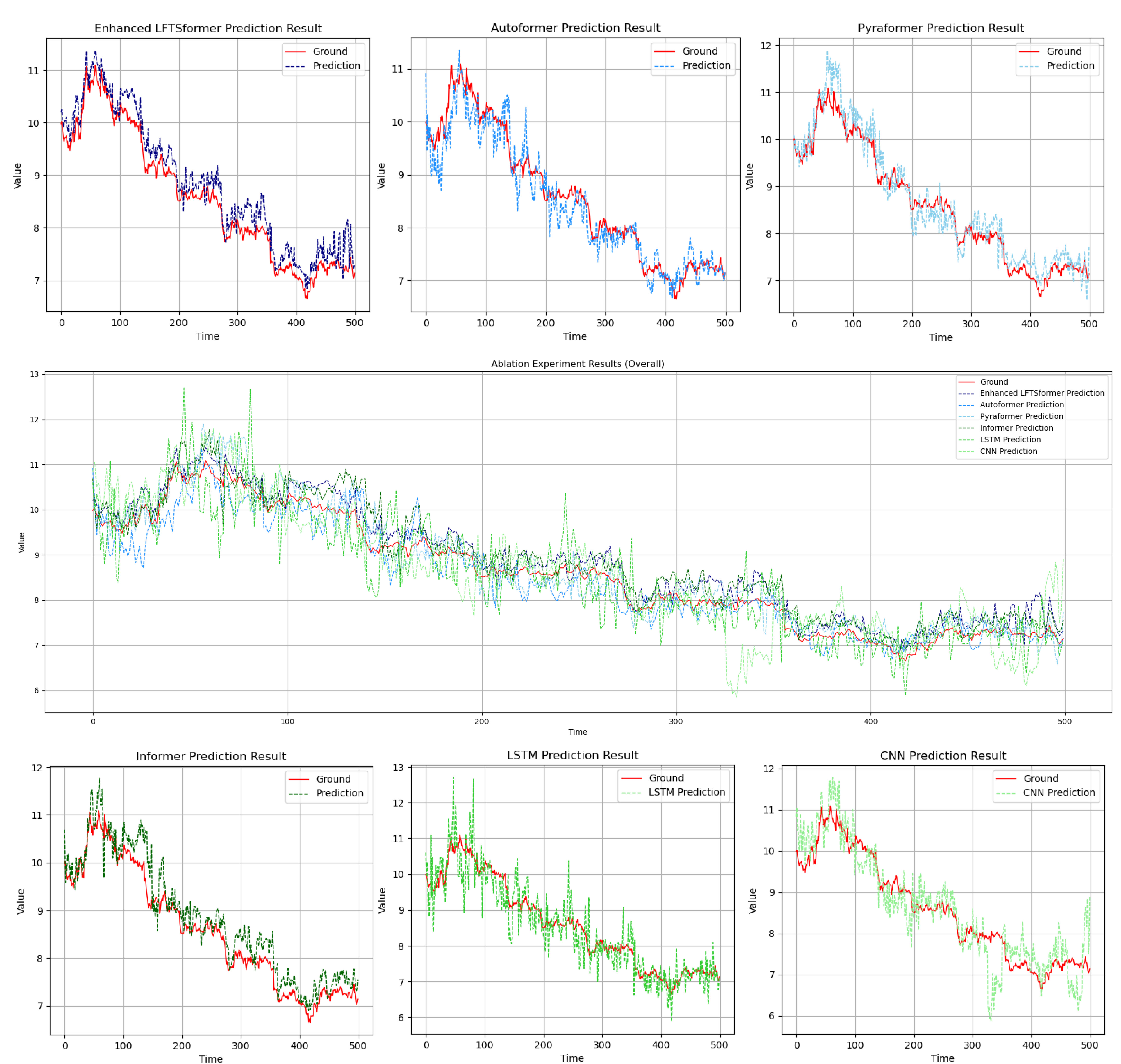}
  \xdef\xfigwd{\the\wd\figbox}
  \usebox\figbox
  \caption{Results of Comparison Experiments for the Whole Test Set}
  \label{Comp}
\end{figure*}
\begin{enumerate}
\item \textbf{Outstanding Performance}:
Across all metrics, the Enhanced LFTSformer demonstrates improvements over traditional models and basic informer constructs in long-term financial time series forecasting. Although on certain metrics such as MAE/MSE, the performance of Enhanced LFTSformer is marginally lower compared to Pyraformer and Autoformer, the blended heatmap analysis indicates that the Enhanced LFTSformer exhibits deeper colors, signifying superior outcomes within a more holistic evaluation framework.

\item \textbf{Decomposition Capability}:
Curve analysis under similar data processing conditions reveals that the VMD algorithm exhibits enhanced data decomposition capabilities over conventional algorithms like EMD. This is evidenced by its effectiveness in attenuating non-smooth features within the original data, thus improving the accuracy of stock predictions.

\item \textbf{Long-Term Predictive Proficiency}:
Further data examination indicates divergences in curve fitting among different models as the temporal scale of prediction extends. While most curves initially maintain a high degree of fit, they begin to deviate from actual values over time. However, the Enhanced LFTSformer maintains its predictive accuracy, underscoring its capacity to address the challenges of long-term sequential forecasting.
\end{enumerate}

In conclusion, Enhanced LFTSformer represents a superior predictive combination model, holding significant practical implications, especially in domains like long-term investment decisions, risk management, and asset allocation.

\subsubsection{Stability Experiment}
Previous experimental outcomes have established that the Enhanced LFTSformer delivers superior performance on the SHA: 600000 dataset compared to other benchmark models, demonstrating its formidable capacity for long-term sequence prediction. Nevertheless, stock prices, as intricately complex financial time series indices, exhibit unpredictability due to shifts in industrial domains, unforeseen events, fluctuations in market sentiment, and the interplay of chaotic factors.

\begin{figure*}[t!]
  \centering
  \includegraphics[width=\textwidth]{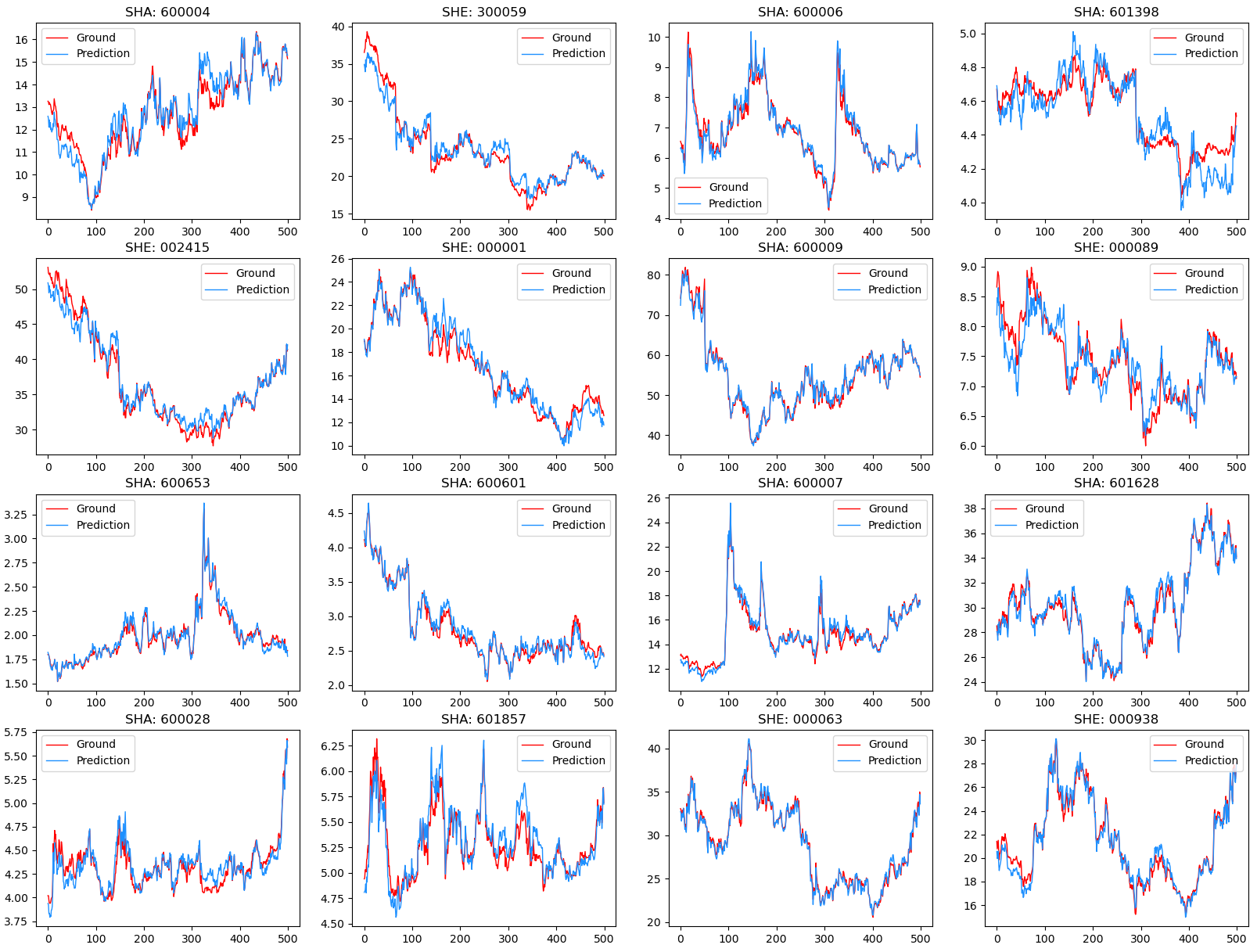}
  \xdef\xfigwd{\the\wd\figbox}
  \usebox\figbox
  \caption{Results of Stability experiments for the whole test set}
  \label{Stable}
\end{figure*}

\begin{figure*}[t!]
  \centering
  \includegraphics[width=0.87\textwidth]{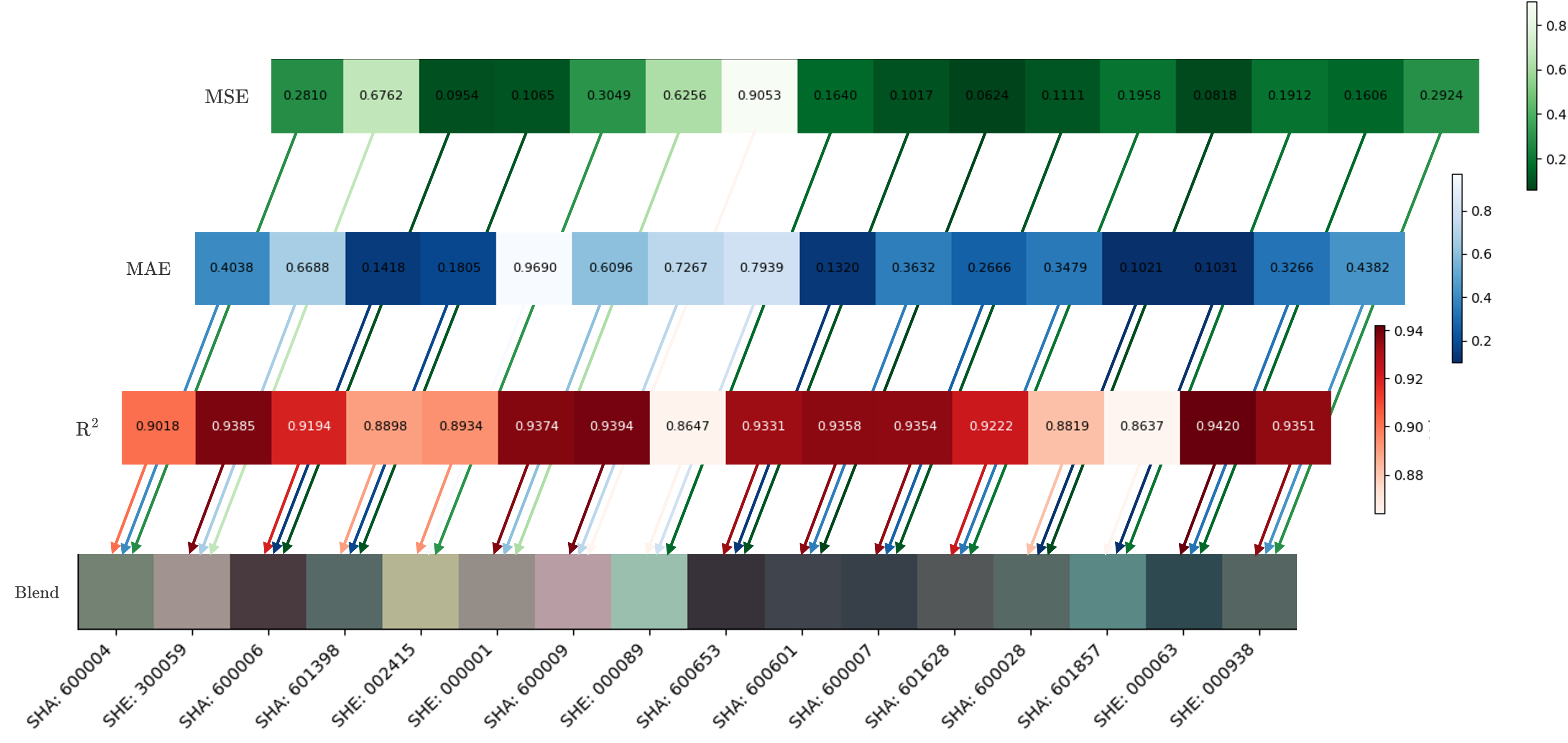}
  \xdef\xfigwd{\the\wd\figbox}
  \usebox\figbox
  \caption{Heat Map for MAE/MSE, $\text{R}^{2}$ and Their Mix for Different Stocks in Stability Experiments.  (Darker colours in the blend heat map represent a better combination of predicted performance)}
  \label{HS}
\end{figure*}

To further ascertain the model's stability and adaptability, it was retrained and its predictions reassessed on an additional 21 stocks listed on the Shenzhen or Shanghai Stock Exchanges, with all training configurations and parameters remaining consistent. The results are illustrated in Fig\ref{Stable} (due to space constraints, only 16 out of 21 are randomly showcased). Error heatmaps and blended heatmaps are displayed in Fig\ref{HS}.

Analysis of the tabulated results indicates that the Enhanced LFTSformer achieved commendable predictive outcomes across various stocks in different industry sectors. Nonetheless, it is imperative to acknowledge that our model exhibited significant predictive performance declines in certain stocks, such as SHE: 000089 (Shenzhen Airport) and SHA: 600009 (Shanghai Airport), which are characterized by market capitalizations not exceeding 100 billion RMB and are more susceptible to singular factors.

\section{Conclusions \& Future Work} \label{5}
\subsection{Conclusions}

The operation of stock markets is influenced by a plethora of factors, including economic conditions, seasonal changes, and the global financial climate. These factors may introduce a multitude of anomalies and non-smooth characteristics into the data of stock markets, posing substantial challenges to improving the accuracy and performance of stock price forecasts. Consequently, this paper proposes an adaptive hybrid model for stock prediction, leveraging an enhanced VMD, Feature Engineering (FE), Stacked \& Distributed (DS) Informer, and an adaptive loss function. The Enhanced LFTSformer model, as a promising hybrid model, is capable of adaptively predicting inherently volatile stock data. The primary advantages of this methodology are summarized as follows:

\begin{enumerate}
    \item \textbf{Hybrid Integration}: The Enhanced LFTSformer represents a comprehensive model that integrates advanced feature engineering, an improved encoder in the Informer, and GC-Enhanced optimization. It amalgamates the strengths of various techniques, delivering superior accuracy and robustness. This superiority is especially prominent in long-term time series forecasting.
    
    \item \textbf{Data Decomposition}: Compared to conventional decomposition methods, the MIC-enhanced VMD excels at extracting non-linear features from raw data, effectively augmenting data quality and streamlining the forecasting process.
    
    \item \textbf{Adaptive Response}: A comprehensive analysis of experimental results illustrates the adaptive loss function's rapid response to volatile financial time series data, skillfully managing anomalies and predicting evolutionary trends.
    
    \item \textbf{Generalization Capability}: Experiments on additional stock datasets highlight the exemplary predictive performance of our proposed model, confirming the Enhanced LFTSformer's significant generalization capability and its promising prospects in stock data forecasting.
\end{enumerate}

In conclusion, hybrid stock prediction models that integrate the advantages of multiple algorithms demonstrate increased predictive accuracy and robustness. Notably, their effectiveness is sustained in long-term time series forecasting.

\subsection{Future Work}
While this study has made preliminary advancements, several areas warrant further investigation to deepen our understanding and enhance the practical applications of stock market prediction models. The following research directions are proposed:

\begin{enumerate}
    \item \textbf{Predictive Modeling for Small Enterprises}: A notable limitation of this study is its focus predominantly on large-scale enterprise stocks. The stock prices of smaller firms, however, may exhibit greater sensitivity and vulnerability to unforeseen events. Future research should explore the integration of natural language processing and real-time event analysis to more accurately forecast stock price fluctuations of these smaller entities. It is crucial to develop specific parameters or variables that can effectively capture the nonlinear impacts of such events on stock prices.

    \item \textbf{Non-linear Optimization in Feature Engineering}: Another limitation identified in this study is the primarily linear methodology employed in our feature engineering. Future research could investigate non-linear techniques, such as activation functions in neural networks or dynamically adjusted weight coefficients, to perpetually optimize the generation of feature values during iterative learning. Such approaches are anticipated to enhance model performance and adaptability.

    \item \textbf{Further Optimizations and Refinements}: This study has highlighted two main areas for potential enhancement. First, our feature selection was limited to correlation factors, overlooking important metrics such as the F-score and sensitivity. Subsequent studies should incorporate these metrics to evaluate the relationships between variables and stock characteristics more thoroughly, aiming to reduce data redundancy. Second, the selection of parameters in our models may lack precision and comprehensiveness. Future studies should consider the application of optimization algorithms to address the sensitivity issues inherent in the parameter selection of deep learning networks.
\end{enumerate}

By exploring these proposed avenues, we anticipate further improvements in the precision, utility, and comprehensiveness of stock market prediction models. These advancements will not only propel academic research in the financial domain but also provide practitioners with more accurate decision-support tools, facilitating navigation through the complex challenges of financial markets.

\begin{IEEEbiography}[{\includegraphics[width=1in,height=1.25in,clip,keepaspectratio]{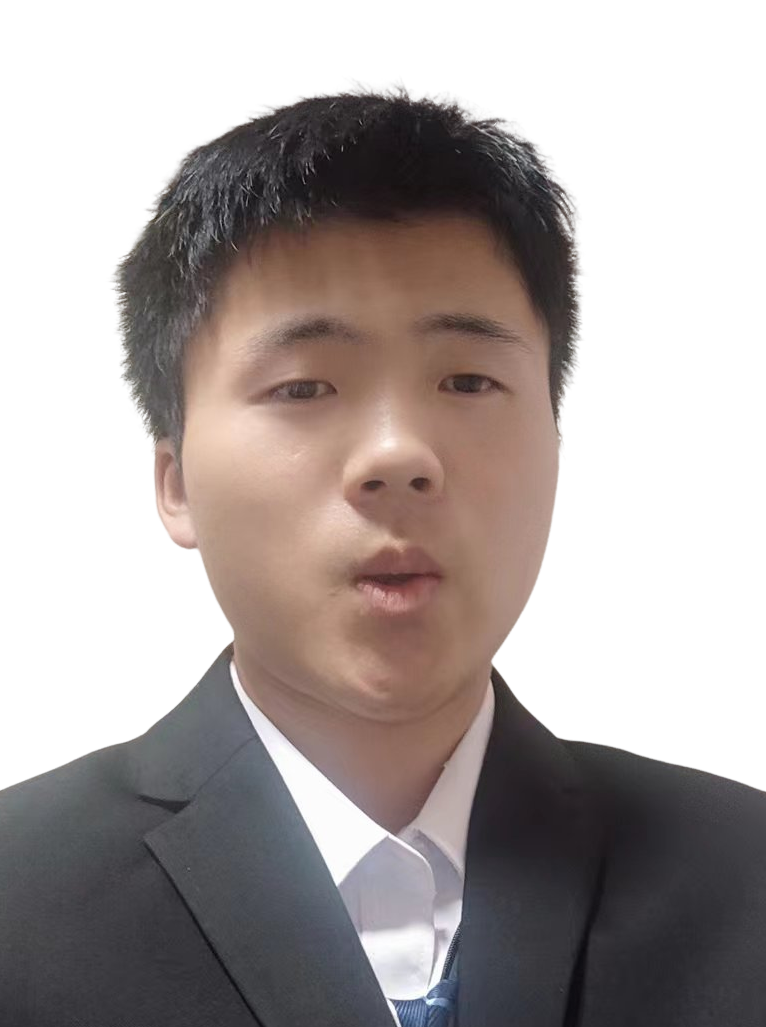}}]{Jian'an Zhang}
(M'2021) Zhang Jian'an was born on June 21, 2003, in Pingxiang, Jiangxi Province, China. He is currently pursuing a Bachelor's degree at the School of Mathematics, Shanghai University of Finance and Economics, with a minor in Finance at Fudan University. Since 2021, he has been a student member of IEEE. He has presented papers at several international computer conferences and has served as a presenter at ICAACE, as well as a reviewer for related papers. His primary research focuses on using deep learning techniques to solve practical problems such as time series analysis and option pricing
\end{IEEEbiography}
\begin{IEEEbiography}
[{\includegraphics[width=1in,height=1.25in,clip,keepaspectratio]{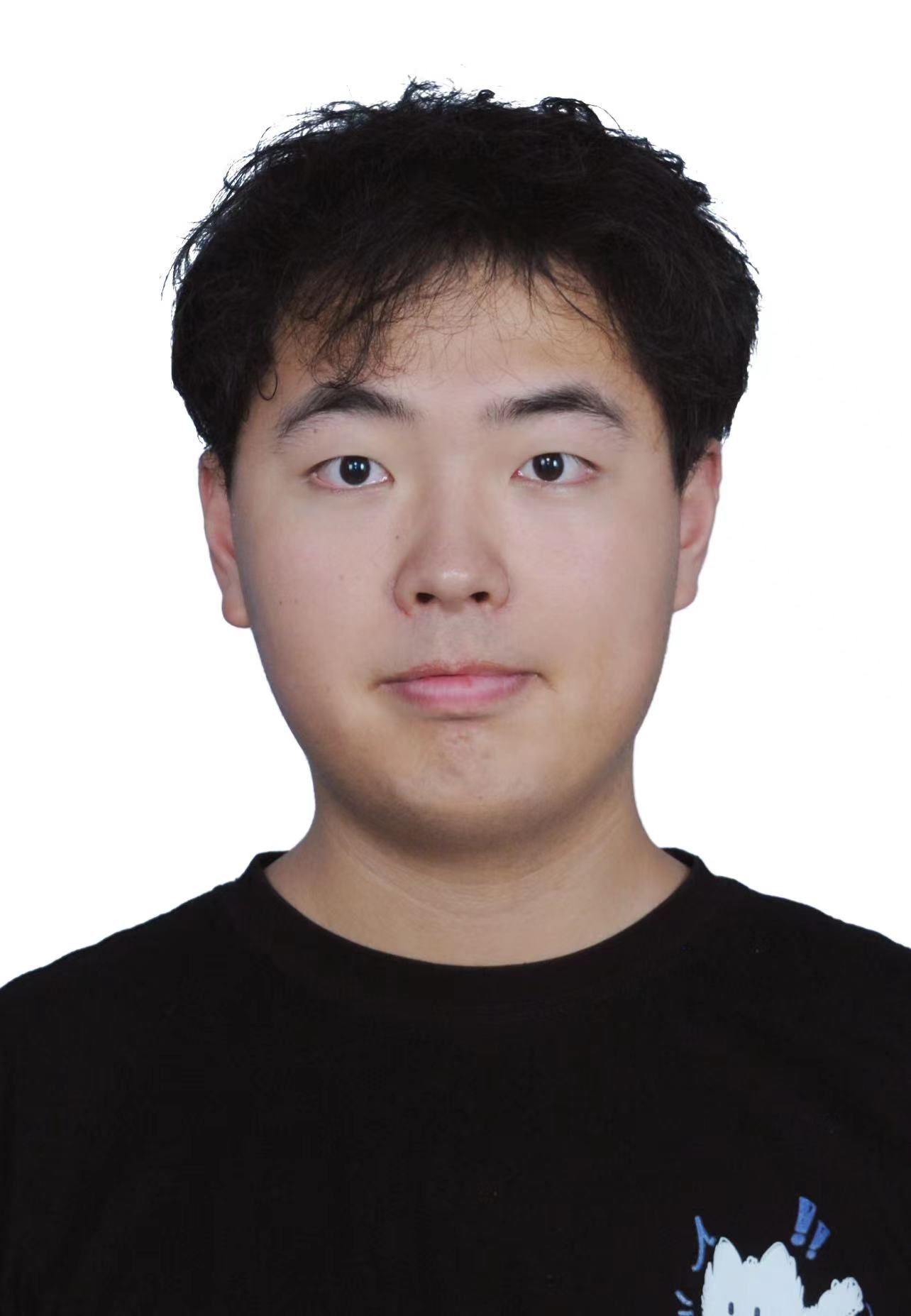}}]{Hongyi Duan}
(M'2023) Hongyi Duan was born in Shanghai, China, on September 2, 2003. He is currently pursuing the Bachelor’s degree in computer science and artificial intelligence at Xi'an Jiaotong University, Xi'an, Shaanxi, People's Republic of China, expected in 2025. Mr. Duan has been an IEEE Student Member since 2023. 

Despite being early in his academic career, Mr. Duan has already amassed numerous research achievements. Among his numerous publications, two are particularly representative: "Application and Analysis of Machine Learning Based Rainfall Prediction," presented at the 2023 8th International Conference on Intelligent Computing and Signal Processing (ICSP), doi: 10.1109/ICSP58490.2023.10248891; and "Comparative study of microgrid optimal scheduling under multi-optimization algorithm fusion," presented at the 2023 10th International Forum on Electrical Engineering and Automation (IFEEA), doi: 10.1109/IFEEA60725.2023.10429466.
\end{IEEEbiography}

\EOD

\end{document}